\documentclass[journal]{IEEEtran}

\usepackage{cite}

\usepackage{textcomp}

\usepackage{amsmath}
\usepackage[linesnumbered,ruled]{algorithm2e}

\usepackage[usenames,dvipsnames]{xcolor}
\usepackage{tcolorbox}
\usepackage{tabularx}
\usepackage{array}
\usepackage{colortbl}
\tcbuselibrary{skins}
\newcolumntype{Y}{>{\raggedleft\arraybackslash}X}

\tcbset{tab1/.style={fonttitle=\bfseries\large,fontupper=\normalsize\sffamily,
colback=yellow!10!white,colframe=red!75!black,colbacktitle=Salmon!40!white,
coltitle=black,center title,freelance,frame code={
\foreach \n in {north east,north west,south east,south west}
{\path [fill=red!75!black] (interior.\n) circle (3mm); };},}}

\tcbset{tab2/.style={enhanced,fonttitle=\bfseries,fontupper=\normalsize\sffamily,
colback=yellow!10!white,colframe=red!50!black,colbacktitle=Salmon!40!white,
coltitle=black,center title}}

\ifCLASSINFOpdf
\else
\fi
\usepackage{algorithmic}

\usepackage{array}

\ifCLASSOPTIONcompsoc
  \usepackage[caption=false,font=normalsize,labelfont=sf,textfont=sf]{subfig}
\else
  \usepackage[caption=false,font=footnotesize]{subfig}
\fi

\usepackage{stfloats}

\usepackage[utf8]{inputenc} 
\usepackage[T1]{fontenc}    
\usepackage{hyperref}       
\usepackage{url}            
\usepackage{booktabs}       
\usepackage{amsfonts}       
\usepackage{nicefrac}       
\usepackage{microtype}      
\usepackage{booktabs}
\usepackage{multirow}
\usepackage{siunitx}
\usepackage{graphicx}
\usepackage{amsmath}
\usepackage{lmodern}
\usepackage{etoolbox}

\begin{document}
%
\title{MAMMO: A Deep Learning Solution for \\Facilitating Radiologist-Machine Collaboration in Breast Cancer Diagnosis}
%
%
%

\author{Trent~Kyono,
        Fiona J. Gilbert,
        Mihaela van der Schaar,~\IEEEmembership{Fellow,~IEEE}
\thanks{T. Kyono and M. van der Schaar are with the Department
of Computer Science, University of California, Los Angeles, CA 90095, USA,
e-mail: (tmkyono@ucla.edu).}
\thanks{F. Gilbert is with the Department of Radiology, Cambridge Biomedical Research Campus, University of Cambridge.}%
}%

\maketitle

\begin{abstract}

With an aging and growing population, the number of women requiring either screening or symptomatic mammograms is increasing. To reduce the number of mammograms that need to be read by a radiologist while keeping the diagnostic accuracy the same or better than current clinical practice, we develop Man and Machine Mammography Oracle (MAMMO) - a clinical decision support system capable of triaging mammograms into those that can be confidently classified by a machine and those that cannot be, thus requiring the reading of a radiologist.
The first component of MAMMO is a novel multi-view convolutional neural network (CNN) with multi-task learning (MTL). MTL enables the CNN to learn the radiological assessments known to be associated with cancer, such as breast density, conspicuity, suspicion, etc., in addition to learning the primary task of cancer diagnosis. We show that MTL has two advantages: 1) learning refined feature representations associated with cancer improves the classification performance of the diagnosis task and 2) issuing radiological assessments provides an additional layer of model interpretability that a radiologist can use to debug and scrutinize the diagnoses provided by the CNN. The second component of MAMMO is a triage network, which takes as input the radiological assessment and diagnostic predictions of the first network's MTL outputs and determines which mammograms can be correctly and confidently diagnosed by the CNN and which mammograms cannot, thus needing to be read by a radiologist. Results obtained on a private dataset of 8,162 patients show that MAMMO reduced the number of radiologist readings by 42.8\% while improving the overall diagnostic accuracy in comparison to readings done by radiologists alone. We analyze the triage of patients decided by MAMMO to gain a better understanding of what unique mammogram characteristics require radiologists' expertise.

\end{abstract}


%
\IEEEpeerreviewmaketitle

\section{Introduction}
\IEEEPARstart{B}{reast} cancer is the most prevalent cancer diagnosed in women, with nearly one in eight women developing breast cancer at some point in their lifetime.  
With the inclusion of screening mammography into breast cancer prevention and detection, randomized clinical trials have shown a 30\% reduction of breast cancer mortality in asymptomatic women \cite{duffy-2002}.  The success of these early breast cancer screening programs has lead to an increase in the total number of annual mammography exams conducted in the US alone to nearly 40 million \cite{broeders-2012}.  Screening mammograms are usually read by two highly trained specialists, which is costly and can be prone to variation and error \cite{kooi-2017}. Many recent research efforts \cite{ akselrod-2016,huynh-2016, qiu-2016,  samala-2016b,  abbas-2016, jiao-2016, shen-2017,becker-2016,kooi-2017,akselrod-2017,carneiro-2017,ribli-2017, mohamed-2018} have been motivated by the increasing number of mammograms requiring reading, presenting an opportunity to automate and reduce the additional workload and responsibility placed on radiologists.

The recent success of convolutional neural networks (CNNs) in computer vision tasks has resulted in an influx of publications and implementations applying CNNs to mammography.  A recent publication by \cite{karssemeijer-deeplearning} showed radiologists significantly improved performance when using a deep learning computer system as decision support in breast cancer diagnosis.  There are two primary objectives or themes in the existing literature applying deep learning to mammography.  The first, which occupies the majority of the research share, is to assist the radiologists in making decisions through computer-aided detection (CAD) as in the works of \cite{jamieson-2012, akselrod-2016,huynh-2016, kooi-2016, qiu-2016, samala-2016a, samala-2016b, dheeba-2014, abbas-2016, jiao-2016}.  The objective here is to assist the radiologist with diagnostic decisions, but does not allow for any patient to bypass radiologist reading.    The second objective, which has recently gained popularity, involves training CNNs to diagnose a patient without radiologist reading \cite{shen-2017,becker-2016,kooi-2017,akselrod-2017,carneiro-2017,ribli-2017, mohamed-2018}. Although results are promising, these methods do not provide a solution for clinical integration other than replacing radiologists entirely or as a second opinion.

\begin{figure}[t]
  \includegraphics[width=\linewidth]{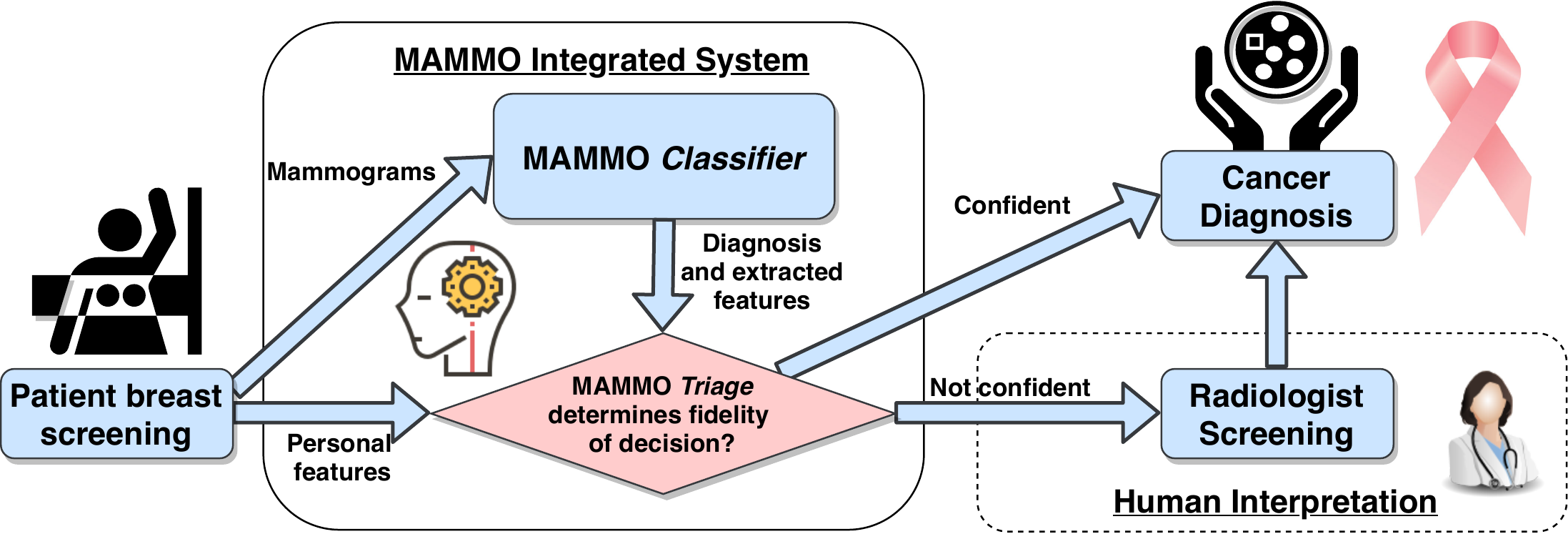}
  \caption{Flow diagram showing integration of MAMMO into the operational clinical setting.  Radiologist reading is circumvented whenever MAMMO is confident in the \textit{Classifier}'s predictive accuracy, i.e., MAMMO \textit{Triage} classifies the patient as one that \textit{Classifier} will diagnose accurately.}
  \label{fig:mammo}
\end{figure}

In this paper we present Man and Machine Mammography Oracle (MAMMO): a clinical decision support system (CDSS) that is capable of learning and reducing the number of patients requiring radiological reading.  As shown in Fig.~\ref{fig:mammo}, MAMMO is comprised of two components: 1) MAMMO \textit{Classifier}, a multi-view CNN trained using multi-task learning (MTL) that enables the CNN to learn the radiological assessments known to be associated with cancer, such as breast density, conspicuity, suspicion, etc., and 2) MAMMO \textit{Triage}, which takes as input the radiological assessment and diagnostic predictions of MAMMO \textit{Classifier} and determines which mammograms can be correctly and confidently diagnosed from those that cannot, thereby requiring reading by a radiologist.  On the basis of this learning, we discover that patients sent to the radiologist by MAMMO \textit{Triage} are those with attributes known to be associated with breast cancer, such as high breast density, older age, etc.  As a result, MAMMO will provide radiologists with more time to focus their attention on the difficult or complex cases that MAMMO is not confident in diagnosing accurately, providing immediate gains in terms of time and cost savings that are unparalleled in the current literature.

We briefly introduce some of the related works as it pertains to this work and provide a more detailed description in Appendix~\ref{Appendix:related}.

\subsection{CAD limitations}
CAD for mammography has been around since the 1990s and relies on conventional computer vision techniques centralized around the detection of hand-crafted imaging features \cite{jamieson-2012}.  Although incremental software updates have been made to improve the detection rates of CAD over the years, this trend has waned over the past decade.  Initial studies on CAD efficacy have shown improvements in breast cancer detection \cite{dheeba-2014}.  Later, larger and more comprehensive studies have disputed these claims.  It has been shown in one of the most definitive studies of over 500,000 mammograms in \cite{lehman-2015} that CAD does not statistically help classification performance of mammograms across any population due to dissonance between radiologist and CAD, i.e. many radiologists either rely only on CAD output or fail to use it all together.  Additionally, CAD increases the average time to evaluate a patient by 20\% due to the additional software interfacing required \cite{cad-takesmoretime}.  

A majority of the current machine learning literature for CAD is centered around using CNNs for improving the detection rates of malignant soft tissue (masses) or micro-calcifications, or alternatively, tasks such as density classification.  Although these works show improvement over existing hand-crafted feature-based methods, it does not address the potential adverse effects of CAD on radiologists performance in operation and does not explore the potential for machine only reading in mammograms.  This still leaves the radiologist with the same number of patients to read. One approach has been to use CAD as an independent second reader with a consensus read of the recalled cases \cite{nishikawa-2018}.  

\subsection{Radiologist-machine collaboration}

The research involving CNNs and mammography are centered around detection, classification, or both.  The most popular task in this domain is to diagnose cancer or predict BI-RADS (\textit{Breast Imaging Report and Data System}) and benchmark performance against that of a radiologist.  There are two common directions that CNNs have been used for diagnosing breast cancer.  The first method, utilizes region-of-interests (ROIs), such as patch-based or sliding window detection, region proposal networks, one-shot (or two-shot) detectors, etc., and has the highest reported accuracy for cancer detection, often surpassing radiologist capability.  However, these works rely on a very scarce commodity: a dataset annotated with benign or malignant locations.  Because of this, the same \textit{limited} public datasets, DDSM (\textit{Digital Database for Screening Mammography}) or INbreast \cite{ddsm-database1}, are almost exclusively used throughout the literature of ROI methods.  Though these methods are exceptional at detecting very low-level features, they typically fail when a diagnosis requires knowledge of high level contextual features that a radiologist would look for, such as the symmetry between the left and right breast or subtle changes compared to a prior examination \cite{tommy-2015}.  The second alternative is image-level classification where a network is trained without requiring ROI.  All of the current image-level publications \cite{krysztof-etal-2017, carneiro-2015} report classification capabilities less than an adept radiologist.  Although image-level classification requires significantly more training data \cite{end-to-end-nips}, it has the potential to surpass ROI methods since it is not dependent on costly annotated locations and learns from the full mammogram in a true data driven fashion.    This work focuses on \textit{image-level} classification and does not use ROI. 

\subsection{Contributions}

MAMMO is a clinical decision support system that is capable of reducing the number of patients requiring mammogram reading by a radiologist. MAMMO is capable of autonomously diagnosing a patient and providing a recommendation of whether this diagnosis is valid or will require additional radiologist scrutinization.  
The design of MAMMO is grounded to a new application for supervised learning that provides the following technical contributions:

\begin{itemize}
    \item \textit{Workload reduction:} MAMMO provides the first solution for reducing the overall number of mammograms a radiologist would need to read by learning to distinguish which patient attributes contribute to a confident machine learning prediction from those that do not.  Experiments show that MAMMO learns to screen the patients with attributes that are generally associated with lower risk, such as patients with no family history of breast cancer, younger patients, patients with lower breast densities, etc., leaving the radiologist with more time to focus on the more complex cases that require meticulous reading.  
    \item \textit{Multi-task learning:} The construction of MAMMO \textit{Classifier} is the first to leverage MTL in image-level mammogram diagnosis providing two important benefits: 1) improving the predictive performance of CNN approaches, and 2) improving the interpretability and usability of AI approaches by predicting both malignancy and radiological labels known to be associated with cancer (e.g. conspicuity, suspicion etc.) that a radiologist could debug and question to better interact with and interpret MAMMO issued predictions.
    
    \item \textit{Challenging dataset:} In comparison to other publications that use publicly available datasets, results are presented on the most challenging dataset to date, comprised mainly of cases recalled from screening.  The dataset, which will be discussed later, was designed to challenge a human reader and contains a high concentration (estimated to be greater than 50\% \cite{tommy-2015}) of patients with overlapping tissues on their mammograms that falsely manifest themselves as suspicious features.  

\end{itemize}

%
%
%
%


\section{MAMMO formalization} 

\begin{figure}[t!]
\centering
    \includegraphics[width=\linewidth]{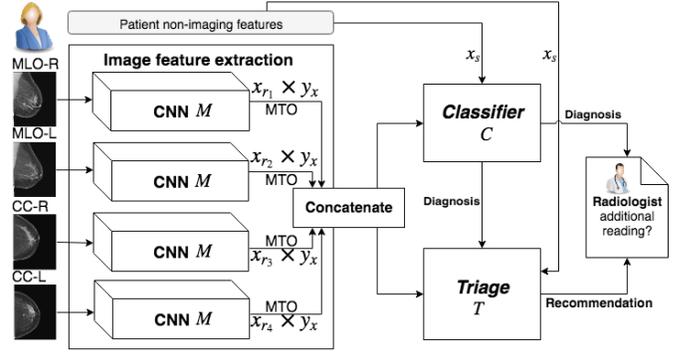}
    \caption{\label{fig:mammocomponents} System-level illustration of MAMMO framework. MTO stands for multi-task outputs.}
\end{figure}

MAMMO uses supervised learning to distinguish which mammograms can safely be screened by machine learning models without radiologist interpretation from those that need further scrutinization.  In this section, MAMMO is formalized according to the illustration in Fig.~\ref{fig:mammocomponents}. The system performs two primary predictions: 1) breast cancer diagnosis at MAMMO \textit{Classifier}, and 2)  fidelity evaluation of \textit{Classifier} predictions at MAMMO \textit{Triage}.

Let  $\mathcal{X} = \mathcal{X}_s \times \mathcal{X}_m$, $\mathcal{X}_r$, and $\mathcal{Y}$  be three spaces, where  $\mathcal{X}_s$ is the patients' non-imaging feature space (such as age), $\mathcal{X}_m$ is the patients' mammogram imaging feature space, $\mathcal{X}_r$ represents the radiologists interpreted mammogram features (such as breast density, conspicuity, etc.), and $\mathcal{Y}$ is the space of all possible diagnoses, that is $\mathcal{Y} = \{0,1\}$, where 0 corresponds to normal and 1 corresponds to malignancy.   

Given a patient, $x \in \mathcal{X}$, let a radiologist as a classifier be defined as a map, $R: \mathcal{X} \rightarrow \mathcal{X}_r \times \mathcal{Y}$, which takes as input a patient's non-imaging features, $x_s \in \mathcal{X}_s$, and mammograms, $x_m \in \mathcal{X}_m$. $R$ provides as output the radiological annotation, $x_r \in \mathcal{X}_r$, and the patient's cancer outcome, $y_x \in \mathcal{Y}$.  
For patient $x$ with mammogram views $x_m \in \mathcal{X}_m = \mathcal{X}_{m_1} \times \mathcal{X}_{m_2} \times \mathcal{X}_{m_3} \times \mathcal{X}_{m_4}$, let $\mathcal{X}_{m_i}$ represent a view from a patient's four mammogram views: medioloateral oblique (MLO) right and left, and craniocaudal (CC) right and left.   Additionally, for each of $x$'s mammogram views, $x_{m_i} \in \mathcal{X}_{m_i}$, the radiologist prediction for that $i$-th view is $x_{r_i} \in \mathcal{X}_{r_i}$, such that $x_r \in \mathcal{X}_r =  \mathcal{X}_{r_1} \times \mathcal{X}_{r_2} \times \mathcal{X}_{r_3} \times \mathcal{X}_{r_4}$.
MAMMO CNN is defined by a map, $M: \mathcal{X}_{m_i} \rightarrow \mathcal{X}_{r_i}  \times \mathcal{Y}$, where $M$ takes as input one of a patient's mammogram views, $x_{m_i} \in \mathcal{X}_{m_i}$, and outputs the radiologist prediction for that view, $x_{r_i} \in \mathcal{X}_{r_i}$,  and the patient's actual cancer outcome, $y_x \in \mathcal{Y}$. \textit{Classifier} is defined as a map, $C :  \mathcal{X}_s \times \mathcal{X}_r \rightarrow \mathcal{Y}$, where $C$ takes as input the patient's non-imaging features, $x_s \in \mathcal{X}_s$, and the CNN predicted radiologist features, $M(x_{m_1}) \times M(x_{m_2}) \times M(x_{m_3}) \times M(x_{m_4}) \in \mathcal{X}_r$. $C$ outputs a diagnostic prediction of the actual cancer outcome, $y_x \in \mathcal{Y}$.
Similarly, \textit{Triage} is defined as a map, $T : \mathcal{X}_s \times \mathcal{X}_r \times \mathcal{Y} \to \{0, 1\}$, where $T$ takes as input the patient's non-imaging features, $x_s \in \mathcal{X}_s$, the CNN predicted radiologist features,  $M(x_{m_1}) \times M(x_{m_2}) \times M(x_{m_3}) \times M(x_{m_4}) \in \mathcal{X}_r$, and the classifiers prediction, $C(x_s, M(x_{m_1}), M(x_{m_2}), M(x_{m_3}), M(x_{m_4}) ) \in \mathcal{Y}$. $T$ outputs $1$ when the radiologist is not needed and \textit{Triage} is confident in the prediction of Classifier, and $0$ otherwise.

Generating $T$ requires optimizing the agreement between $C$ and $R$ with the actual outcome $\mathcal{Y}$. 
Let the expected false positive rate and expected false negative rate for $x$ be $\mathbb{E}[$FPR$]$ and $\mathbb{E}[$FNR$]$, respectively.  The primary design goal of $T$ is to reduce the number of mammograms $R$ reads by offloading patients to $C$, without increasing the false negative rate ($\text{FNR}_{R}$) and false positive rate  ($\text{FPR}_{R}$) that $R$ would have performed at if $R$ were to have read all the patients. An optimal and desirable model for patient triage is one that minimizes the probability, $w_x$, that a patient, $x$, needs their mammograms read by a radiologist, which is solved by the following constrained optimization:

\begin{equation} \label{eq:maximize}
   \begin{array}{rrclcl}
    \displaystyle \min_{x \in \mathcal{X}} & \multicolumn{3}{l}{\mathbb{E}[w_x]} \\
    \textrm{s.t.} & \text{$\mathbb{E}[$FNR$]$} \leq \text{FNR}_{R} \text{, and}\\
     & \multicolumn{3}{l}{\text{$\mathbb{E}[$FPR$]$} \leq \text{FPR}_{R}} \\
\end{array}
\end{equation}

\begin{figure*}[!t]
  \includegraphics[width=\linewidth]{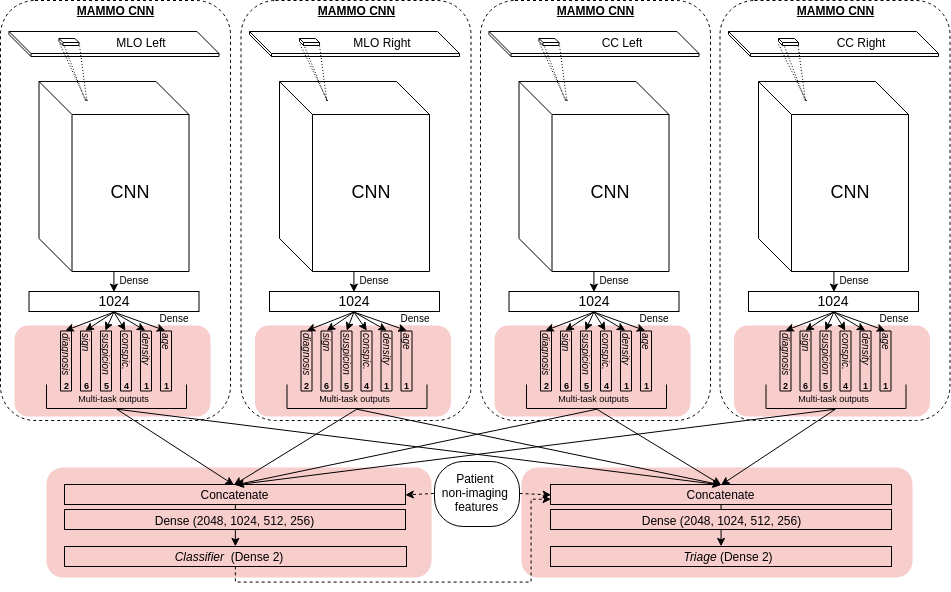}
  \caption{Full MAMMO network as a stacked classifier of 4 MAMMO CNNs. Highlighted are the multi-task outputs of each view and how they are fused to generate the \textit{Classifier} and \textit{Triage} networks.}
  \label{fig:mammocnn}
\end{figure*}

\section{MAMMO System}

This section discusses the two main functional components of MAMMO: the \textit{Classifier} and \textit{Triage}.  The neural network architecture for the the overall system is shown in Fig.~\ref{fig:mammocnn} 

\subsection{MAMMO Classifier}
MAMMO \textit{Classifier}, $C$, is a multi-view CNN designed to provide an accurate diagnosis given a patient's four mammogram views. Consider Fig.~\ref{fig:mammocomponents}, $C$ is trained in two consecutive stages.  
The primary objective of the first training phase is to generate a CNN, $M$, that predicts both the diagnosis and radiological assessments on an individual mammogram view basis. 
The objective of our second training phase is to train a classifier that takes as input the multi-task outputs (MTO) of $M$ for each of a patient’s four mammogram views and predicts a patient-level diagnosis. In the current literature, this is done by combining multiple mammogram views at the dense layers before the final output layer as done in \cite{krysztof-etal-2017, carneiro-2017}. However, we choose to combine multiple views over the MTO for two reasons.  First, the MTO are extracted imaging features that emulate radiological assessment and are what radiologist would naturally consider when reading multiple mammogram views. For example, breast density asymmetries between left and right breasts are often indicators of cancer \cite{breast-asymmetry1}. Second, since we pre-trained our CNN in the first training phase, the MTO serves as a refined feature space for combining mammograms, requiring no retraining of layers prior to the MTO and improves performance in scenarios where there is limited data.

MTL is used to fine-tune $M$ providing several additional benefits.  The work of \cite{argyriou-2006} demonstrates both empirically and theoretically the performance advantages of learning related tasks simultaneously over each task independently.  This is amplified in situations when some tasks have very few data points and would be nearly impossible to learn individually.  Additionally, MTL is leveraged to learn refined feature representations and improve classification performance of the primary task, diagnosis, by obligating MAMMO CNN to learn the radiological assessment known to be associated with cancer, such as the breast density, conspicuity, or suspicion.  The radiological assessment (MTO) provided by each $M$ are used by $T$ to learn which mammograms the radiologist and $C$ would provide correct or incorrect diagnoses for, thus improving triage performance. 
Finally, concatenating and fusing mammogram views for left and right breast, including corresponding MLO and CC views for each, over the trained MTO provides a reduced (and refined) feature space that improves classification performance, particularly in data-starved scenarios \cite{cascade-ensemble}.  Concatenation of mammogram views could be performed at a subsequent dense layer \cite{stacked-ensemble}, but these layers in practice are typically larger and thus require more training data.  The sources of gain attributed to MTL are shown experimentally in the results section.

\begin{algorithm}[t]
    \SetKwInOut{Input}{Input}
    \SetKwInOut{Output}{Output}

    \underline{function TrainTriage} $(R,C,S)$\;
    \Input{A radiologist classifier $R$, a binary classifier $C$, a set of patients $S$}
    \Output{Optimized \textit{Triage} model $T$ that minimizes the number of patients sent to $R$}
    Partition subsets $S_{train}, S_{val}, S_{test}$ from $S$;\\
    $n_{min} \leftarrow $ number of patients in $S_{val}$;\\
    $FNR_{R}$, $FPR_{R} \leftarrow$ the FNR and FPR of $R$ on $S_{val}$; \\
    \For{$b_R$ and $b_C$ \text{by} $\delta$}{
        $T_n \leftarrow$ train model over $S_{train}$ with loss in Eq.~\ref{eq:loss2}; \\
        \For{$\alpha,\beta \in$ \{all thresholds of $T_n $, $C$\}} {
            $n \leftarrow$ number patients sent to $R$ by $T_n$ on $S_{val}$; \\
            $FNR, FPR \leftarrow$ calculate $\mathbb{E}[$FNR$]$ and $\mathbb{E}[$FPR$]$ from $T_n, R$ and $C$ on $S_{val}$ using $\alpha,\beta$; \\
            \If{$n < n_{min}$, $FNR \leq \text{FNR}_{R}$, $FPR \leq \text{FPR}_{R}$}
            {
                $T, n_{min} \leftarrow T_n, n$; \\
            }
        }
    }
    return $T$;
    \caption{\label{alg0} Trains and returns the best \textit{Triage} model that minimizes the number of patients the radiologist must examine subject to the FNR and FPR constraints in Eq.~\ref{eq:maximize}}
\end{algorithm}

\subsection{MAMMO Triage} 

MAMMO \textit{Triage} reviews the prediction of \textit{Classifier}, while also considering the patient's non-imaging features, such as age, and the radiological predictions of each individual MAMMO CNN, and decides whether or not the diagnosis will be correct or not. 

For a patient, $x \in \mathcal{X}$, with the observed outcome, $y_x \in \mathcal{Y}$, and $R(x)$ is the radiologist's prediction on $x$, the loss attributed to $R(x)$ is $\mathcal{L}_{R}(x, y_x)$ given by 
\begin{equation} \label{eq:lossrad}
   \mathcal{L}_{R}(x, y_x) = \sum_{y_x \neq R(x)}{R(x)}
\end{equation}  
Similarly, the loss attributed to $C(x)$ is $\mathcal{L}_{C}(x, y_x)$ given by
\begin{equation} \label{eq:lossclass}
   \mathcal{L}_{C}(x, y_x) = \sum_{y_x \neq C(x)}{C(x)}.
\end{equation} 
Eq.~\ref{eq:lossrad} and ~\ref{eq:lossclass} account for both the FPR and FNR for $R$ and $C$, respectively.  By the Lagrangian method, the loss function that satisfies Eq.~\ref{eq:maximize} is equivalent to
\begin{equation} \label{eq:loss1}
    \mathcal{L}_{T} = \mathbb{E}[w_x] + \lambda_1 \mathbb{E}[\text{FNR}] + \lambda_2\mathbb{E}[\text{FPR}]\textit{.}
\end{equation}
This is estimated by sample averages as:

\begin{equation} \label{eq:loss2}
    \mathcal{L}_{T} = \sum_{x \in \mathcal{X}}{w_x + b_R w_x\mathcal{L}_{R}(x, y_x) + b_C(1 - w_x)\mathcal{L}_{C}(x, y_x)}\textit{,}
\end{equation}
where $b_R$ and $b_C$ are used for adjusting the triage of patients between $R$ and $C$.  

A detailed explanation of training \textit{Triage} to minimize the number of patients the radiologist must read is provided in Algorithm~\ref{alg0}.  Given a dataset of patients, $S$, Algorithm~\ref{alg0} first partitions $S$ into three disjoint sets $S_{train}$, $S_{val}$, and $S_{test}$, for training, validation and testing, respectively.  $S_{test}$ is held out (and never operated on) to prevent data leakage from training and validation. The minimum number of patients is monitored by the variable, $n_{min}$, which is initialized to the number of patients in $S_{val}$.  $FNR_{R}$ and $FPR_{R}$ are the false negative and false positive rates of $R$ on $S_{val}$, respectively.  Beginning on line 5, Algorithm~\ref{alg0} iterates over all possible tuning factors, $b_R$ and  $b_C$, using an incremental step size of $\delta$, which is experimentally adjusted depending on the training data size and complexity.   On line 6, $T_n$ is the trained model over $S_{train}$ using the loss function defined in Eq.~\ref{eq:loss2}. Lines 7 through 12 iterate over all $\alpha$ and $\beta$ which are the classification thresholds for $T_n$ and $C$, respectively.  On line 9, the false positive and false negative rate are calculated from the overall system comprised of $T_n$, $R$, and $C$ on $S_{val}$ using $\alpha$ and $\beta$.  Lines 10 through 12 ensure that the \textit{Triage} model, $T_n$, that sends the least amount of patients over to $R$ is saved as long as MAMMO adheres to the FPR and FNR constraints on line 10 (corresponding to Eq.~\ref{eq:maximize}).


\section{Data}

\subsection{Tommy dataset}

The \textit{Tommy} dataset was originally compiled to determine the efficacy and diagnostic performance of digital breast tomosynthesis (DBT) in comparison to digital mammography.  The dataset was collected through six NHS Breast Screening Program (NHSBSP) centers throughout the United Kingdom and read by expert radiologists \cite{tommy-2015}.  It is a rich and well-labeled dataset with a total of 8,162 (1,677 malignant) patients, including radiologist predictions and interpretations, density estimates ($\mu = 38.2$, $\sigma=20.7$), age ($\mu = 56.5$, $\sigma=8.75$), pathology outcomes from core biopsy or surgical excision, and both mammography and DBT imaging modalities.  Although not all patients in the \textit{Tommy} dataset underwent biopsy, each patient underwent expert radiological readings of both DBT and mammography modalities that significantly reduce the likelihood of false negative readings \cite{tommy-2015}.  The \textit{Tommy} dataset does not contain ROI annotations, but it does contain many useful radiological assessments that we leveraged for MTL.  

Patient distributions for age, breast density, and dominant radiological features are shown in Table~\ref{table-tommypatients} in Appendix~\ref{Appendix:Tommy}. The \textit{Tommy} dataset was designed to challenge the radiologist with overlapping breast tissue cases.  In this dataset, it is estimated that roughly 50\% of patients have overlapping tissues that show up on standard 2D mammograms that would falsely manifest as suspicious features \cite{tommy-2015}.  
The patient criteria for selection were one of the following: 1) women recalled after routine breast screening between the ages of 47 and 73, or 2) women with a family history of breast cancer attending annual screening between ages of 40 and 49.

\subsection{Mammogram preprocessing and augmentation}
Mammogram processing steps were performed in several stages.  Processed mammograms were converted from DICOM (Digital Imaging and Communication in Medicine) files into uncompressed 16-bit monochrome PNG (Portable Network Graphics) files.  In this step, all mammogram views were rotated and oriented with the breast along the left margin with nipple oriented to the right.  Mammograms were not cropped, and Lanczos down-scaling was used to reduce the full-field mammograms to 320 x 416 pixels, i.e. $\frac{1}{8}$ of the full mammogram height and width of the narrow field mammograms. This maintained and preserved the width-to-height aspect ratio of $1:1.3$ for all mammogram fields of view.  

During training, mammograms were augmented to prevent over-fitting and promote model generalizability.  Image augmentation was run through the \textit{Keras} \cite{keras-2015} image processing generator with random selections from the following pool of augmentations: horizontal and vertical flips, image rotations of up to 20 degrees, image shear of up to 20\%, image zoom of up to 20\%, and width and height shifts of up to 20\%.  The gray-scale augmented mammograms were then stacked into 3 channels and histogram equalized by Contrasted Limited Adaptive Histogram Equalization (CLAHE) with channel stratified clipping and grid sizes as presented in \cite{teare-2017}.    We used the nominal grid sizes and clip limits they presented and enhanced their approach by using it as an additional augmentation. The CLAHE grid size, $g$,  was augmented according to the following equation:
\begin{equation} \label{eq:clahegrid}
a\in\mathcal{U}(-log_2(k), log_2(k)) \mid g(k) = k + a,
\end{equation}
where $k$ is the nominal grid size.  Similarly, the CLAHE clip limit, $c$, was augmented as follows:
\begin{equation} \label{eq:claheclip}
a\in\mathcal{U}(-log_2(l), log_2(l)) \mid c(l) = l + a,
\end{equation}
where $l$ is the nominal clip limit.  After histogram equalization, a Gaussian noise \cite{gaussian-2015} was applied to each color channel with a $\sigma$ of 0.01, followed by image standardization.  When training $\textit{Classifier}$, each of the four input mammograms were augmented with a random set of augmentations drawn from the aforementioned pool of training augmentations.

\section{MAMMO on the Tommy dataset}

MAMMO experiments were conducted on the \textit{Tommy} dataset. Similar to other works in deep learning for mammography that used larger datasets, a held-out test set was used instead of k-fold cross-validation.  1000 randomly selected patients were reserved for a hold-out testing set.  The remaining 7162 patients were randomly partitioned into a MAMMO CNN training set, a \textit{Classifier} training set, a \textit{Triage} training set, and a validation set of 60\%, 15\%, 15\%, and 10\%, respectively.  The additional training sets used for \textit{Classifier} and \textit{Triage}, provided additional samples that the MAMMO CNN had never seen before to promote generalizability \cite{cascade-ensemble, stacked-ensemble}.  For a comprehensive description of architecture and training details refer to Appendix~\ref{Appendix:Architecture}. 

\subsection{MAMMO CNN performance}

A trade study on the CBIS-DDSM dataset was conducted to select the best CNN from a pool of available ImageNet candidates \cite{ddsm-database1, ddsm-database2} and is presented in Appendix~\ref{Appendix:candidate}.  
The best performing network was InceptionResNetV2, and was therefore used as the CNN in this work.  It was instantiated with ImageNet weights and refined using MTL with the tasks shown in Table~\ref{table:multi_task_outs}. The primary output target, \textit{diagnosis}, was one of either malignant or benign (normal) as determined by the outcome of core biopsy.    Five other auxiliary output targets were trained: \textit{sign}, \textit{suspicion}, \textit{conspicuity}, \textit{density}, and \textit{age}.  The \textit{sign}, \textit{suspicion}, and \textit{conspicuity} were categorical output targets representing radiologist interpretation of the observed mammogram.  Both patient \textit{age} and breast \textit{density} were included as auxiliary tasks for improved regularization and for their known correlation with breast cancer \cite{age-breastdensity, age-breastdensity2, wu-etal-2017, karssemeijer-1998, kallenberg-2016, mohamed-2018}. 
Breast \textit{density} was not categorized by the traditional BI-RADS lexicon, but by a percentage density calculated  from a radiologist assessment on a 10-cm VAS (visual analogue scale) as described in \cite{tommy-2015}. For this reason, breast \textit{density} was not learned as a categorical problem but as a regression, hence the normalization.  Table~\ref{table:multi_task_outs} shows the classification and regression performance of each task. The results shown are the average of 100 test-time augmentations (TTA) per sample.   By providing our networks with various ``perspectives'' of the same mammogram, TTA mitigated the likelihood of misinterpreting a solitary sample and significantly improved performance \cite{test-time-augmentation, test-time-augmentation2}. Area under the receiver operating characteristic curve (AUROC) is reported for each categorical task; for regression targets mean absolute error is reported.

\begin{table}[t!]
\caption{ \label{table:multi_task_outs} Multi-task performance for MAMMO CNN by task.}
\centering
{(a) Categorical tasks.}
\begin{tcolorbox}[tab2,tabularx={p{2.2cm}|p{3.6cm}|c}]
    {\normalfont \small \bf \textcolor{red!60!black}{Task}} &
    {\normalfont \small \bf  \textcolor{red!60!black}{Output}} &
    {\normalfont \small \bf \textcolor{red!60!black}{AUROC}} 
    \\ \hline \hline
    {\normalfont \small Diagnosis}   & {\normalfont \small malignant/benign} & {\normalfont \small 0.787}
    \\ \hline \hline
    {\normalfont \small Sign}   & {\normalfont \small none} & {\normalfont \small 0.640}
    \\ \hline
    {\normalfont \small }   & {\normalfont \small circumscribed} & {\normalfont \small 0.641} \\\hline
    {\normalfont \small }   & {\normalfont \small spiculated} & {\normalfont \small 0.810} \\\hline
    {\normalfont \small }   & {\normalfont \small micro-calcification} & {\normalfont \small 0.551} \\\hline
    {\normalfont \small }   & {\normalfont \small distortion} & {\normalfont \small 0.701} \\ \hline
    {\normalfont \small }   & {\normalfont \small asymm. density} & {\normalfont \small 0.581}
    \\ \hline \hline
    {\normalfont \small Suspicion}   & {\normalfont \small normal} & {\normalfont \small 0.627}
    \\ \hline
    {\normalfont \small }   & {\normalfont \small benign} & {\normalfont \small 0.608} \\ \hline
    {\normalfont \small }   & {\normalfont \small probably benign} & {\normalfont \small 0.577} \\ \hline
    {\normalfont \small }   & {\normalfont \small suspicious} & {\normalfont \small 0.681} \\ \hline
    {\normalfont \small }   & {\normalfont \small malignant} & {\normalfont \small 0.815}  
    \\ \hline \hline
    {\normalfont \small Conspicuity}   & {\normalfont \small not visible} & {\normalfont \small 0.625} \\ \hline
    {\normalfont \small }   & {\normalfont \small barely visible} & {\normalfont \small 0.698} \\ \hline
    {\normalfont \small }   & {\normalfont \small visible, not clear} & {\normalfont \small 0.565} \\ \hline
    {\normalfont \small }   & {\normalfont \small clearly visible} & {\normalfont \small 0.634} \\ \hline
\end{tcolorbox}
{(b) Regression tasks, where MAE is mean absolute error.}
\begin{tcolorbox}[tab2,tabularx={l|l|l}]
    {\normalfont \small \bf \textcolor{red!60!black}{Task}} & 
    {\normalfont \small \bf  \textcolor{red!60!black}{Output description}} &
    {\normalfont \small \bf \textcolor{red!60!black}{MAE}}
    \\ \hline \hline
    {\normalfont \small Breast density}   & {\normalfont \small  Radiologist estimated 0-100\%} & {\normalfont \small 14.96\%} \\ \hline 
    {\normalfont \small Age}   & {\normalfont \small Age 40 to 73} & {\normalfont \small 5.97 yrs} \\ \hline
\end{tcolorbox}
\end{table}


\begin{table}[t!] 
\centering
\caption{ \label{table-ranking} Source of gain for MAMMO CNN and \textit{Classifier} (multi-view MAMMO CNN) shown in terms of AUROC and AUPRC against the closest related works of Zhang et. al \cite{fullimage-zhang} and Geras et al. \cite{krysztof-etal-2017} on the \textit{Tommy} dataset.  Mammogram views (MV) is the number of input mammograms used per patient.  MT is checked when multi-tasking was used, otherwise assume single-task.  TA denotes if test-time augmentation was used. }
\begin{tcolorbox}[tab2,tabularx={c|c|c|c|c|c}]
    {\normalfont \scriptsize \bf \textcolor{red!60!black}{Method}} & 
    {\normalfont \scriptsize \bf \textcolor{red!60!black}{MV}} & 
    {\normalfont \scriptsize \bf  \textcolor{red!60!black}{MT}} &
    {\normalfont \scriptsize \bf \textcolor{red!60!black}{TA}} &

    {\normalfont \scriptsize \bf \textcolor{red!60!black}{AUROC}} &
    {\normalfont \scriptsize \bf \textcolor{red!60!black}{AUPRC}}\\ \hline \hline
    {\normalfont \scriptsize Zhang et. al \cite{fullimage-zhang}}  & {\normalfont \scriptsize 1}   && & {\normalfont \scriptsize 0.566} & {\normalfont \scriptsize 0.112} \\ \hline
    {\normalfont \scriptsize MAMMO CNN}  & {\normalfont \scriptsize 1}   && & {\normalfont \scriptsize 0.725} & {\normalfont \scriptsize 0.305} \\ \hline
    {\normalfont \scriptsize MAMMO CNN}  &{\normalfont \scriptsize 1}  & {\normalfont \scriptsize \checkmark}&    & {\normalfont \scriptsize 0.736} & {\normalfont \scriptsize 0.313} \\ \hline
    {\normalfont \scriptsize MAMMO CNN}   &{\normalfont \scriptsize 1}    && {\normalfont \scriptsize \checkmark}  &{\normalfont \scriptsize 0.770} & {\normalfont \scriptsize 0.365} \\ \hline
    {\normalfont \scriptsize MAMMO CNN}  &{\normalfont \scriptsize 1}   & {\normalfont \scriptsize \checkmark}  & {\normalfont \scriptsize \checkmark}      & {\normalfont \scriptsize \textbf{ 0.787}} & {\normalfont \scriptsize \textbf{0.394}} \\ \hline \hline

    {\normalfont \scriptsize Geras et al. \cite{krysztof-etal-2017}}  &{\normalfont \scriptsize 4}   &  & {\normalfont \scriptsize \checkmark}                                   & {\normalfont \scriptsize 0.608} & {\normalfont \scriptsize 0.205} \\ \hline 
    
    {\normalfont \scriptsize MAMMO \textit{Classifier}}  &{\normalfont \scriptsize 4}   &  & & {\normalfont \scriptsize 0.733} & {\normalfont \scriptsize 0.401} \\ \hline 

    {\normalfont \scriptsize MAMMO \textit{Classifier}}  &{\normalfont \scriptsize 4}   &  & {\normalfont \scriptsize \checkmark}                                   & {\normalfont \scriptsize 0.757} & {\normalfont \scriptsize 0.446} \\ \hline 
    
    {\normalfont \scriptsize MAMMO \textit{Classifier}} & {\normalfont \scriptsize 4}   & {\normalfont \scriptsize \checkmark}& {\normalfont \scriptsize}               & {\normalfont \scriptsize 0.754} & {\normalfont \scriptsize 0.460} \\ \hline 
    
    {\normalfont \scriptsize MAMMO \textit{Classifier}} &{\normalfont \scriptsize 4}   & {\normalfont \scriptsize \checkmark}& {\normalfont \scriptsize \checkmark}    & {\normalfont \scriptsize \textbf{0.791}} & {\normalfont \scriptsize \textbf{0.524}} \\ \hline

\end{tcolorbox}
\end{table}

\begin{figure}[t!]
\centering
  \includegraphics[width=0.8\linewidth]{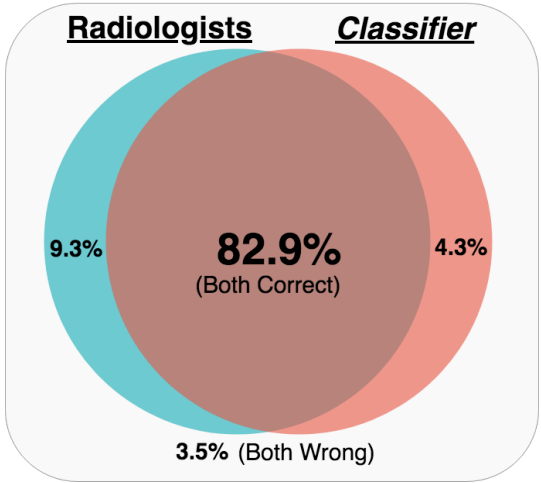}
  \caption{Venn diagram depicting performance of radiologist versus \textit{Classifier}.  Values displayed are the percentage of patients correctly diagnosed by either the radiologists or \textit{Classifier} over the 1000 patient test set. This diagram corresponds directly with the first row in Table~\ref{table-patientdistributions}.}
  \label{fig:venn}
\end{figure}

\begin{table}[htbp] 
\centering
\caption{ \label{table-patientdistributions}Patient distributions over various population strata using a 1000 patient test set.  Values displayed are normalized across each population. $R^+$ and $R^-$ represent radiologists correct and wrong diagnoses, respectively.  Similarly, $C^+$ and $C^-$ represent \textit{Classifier} correct and wrong diagnoses, respectively. }
\begin{tcolorbox}[tab2,tabularx={l|l|l|l|l}]
    {\normalfont \small \bf \textcolor{red!60!black}{Population}} & 
    {\normalfont \small \bf \textcolor{red!60!black}{$R^+C^+$}} &
    {\normalfont \small \bf \textcolor{red!60!black}{$R^+C^-$}} & 
    {\normalfont \small \bf  \textcolor{red!60!black}{$R^-C^+$}} &

    {\normalfont \small \bf \textcolor{red!60!black}{$R^-C^-$}} 
    \\ \hline \hline
    {\normalfont \small All 1000 patients}  & {\normalfont \small 0.829}  & {\normalfont \small 0.093} & {\normalfont \small 0.043} &  {\normalfont \small 0.035}  \\ \hline \hline
    {\normalfont \small High conspicuity}   &{\normalfont \small 0.732} &{\normalfont \small 0.150} & {\normalfont \small 0.068}&  {\normalfont \small 0.050} \\ \hline 
    {\normalfont \small Low conspicuity}   &{\normalfont \small 0.905} &{\normalfont \small 0.048} & {\normalfont \small 0.023}&  {\normalfont \small 0.023} \\ \hline \hline
    {\normalfont \small No sign of cancer}   &{\normalfont \small 0.981} &{\normalfont \small 0.006} & {\normalfont \small 0.000}&  {\normalfont \small 0.013} \\ \hline 
    {\normalfont \small Circumscribed mass}   &{\normalfont \small 0.932} &{\normalfont \small 0.021} & {\normalfont \small 0.021}&  {\normalfont \small 0.025} \\ \hline 
    {\normalfont \small Spiculated mass}   &{\normalfont \small 0.321} &{\normalfont \small 0.543} & {\normalfont \small 0.086}&  {\normalfont \small 0.049}\\ \hline  
    {\normalfont \small Micro-calcification}   &{\normalfont \small 0.647} &{\normalfont \small 0.147} & {\normalfont \small 0.115}&  {\normalfont \small 0.090} \\ \hline  
    {\normalfont \small Distortion}   &{\normalfont \small 0.571} &{\normalfont \small 0.214} & {\normalfont \small 0.167}&  {\normalfont \small 0.048} \\ \hline 
    {\normalfont \small Asymmetrical density}   & {\normalfont \small 0.858} &{\normalfont \small 0.068 } & {\normalfont \small 0.047}& {\normalfont \small 0.026} \\ \hline \hline 
    {\normalfont \small High suspicion ($> 3$)}   &{\normalfont \small 0.235} & {\normalfont \small 0.506} & {\normalfont \small 0.240}&  {\normalfont \small 0.019}\\ \hline 
    {\normalfont \small Low suspicion ($\leq 3$)}   &{\normalfont \small 0.944} &{\normalfont \small 0.013} & {\normalfont \small 0.005}&  {\normalfont \small 0.038} \\ \hline \hline
    {\normalfont \small High breast density}  & {\normalfont \small 0.825} & {\normalfont \small 0.100} & {\normalfont \small 0.037} &{\normalfont \small 0.039} \\ \hline 
    {\normalfont \small Low breast density}   & {\normalfont \small 0.833} &{\normalfont \small 0.086} & {\normalfont \small 0.049}& {\normalfont \small 0.031} \\ \hline  \hline
    {\normalfont \small Family History}   &{\normalfont \small 0.372} &{\normalfont \small 0.402} & {\normalfont \small 0.067}& {\normalfont \small 0.159} \\ \hline
    {\normalfont \small No Family History}  & {\normalfont \small 0.919} &{\normalfont \small 0.032}& {\normalfont \small 0.038} &{\normalfont \small 0.011}\\ \hline\hline 
    {\normalfont \small Age over $\mu$ ($\geq$ 57)}& {\normalfont \small 0.753}   &{\normalfont \small 0.145} & {\normalfont \small .0476} &{\normalfont \small 0.054} \\ \hline
    {\normalfont \small Age under $\mu$ (< 57)} & {\normalfont \small 0.889}  &{\normalfont \small 0.052} & {\normalfont \small 0.039} & {\normalfont \small 0.020} \\ \hline \hline
    {\normalfont \small Recall by 1 reader}   & {\normalfont \small 0.862} &{\normalfont \small 0.092} & {\normalfont \small 0.034}& {\normalfont \small 0.011} \\ \hline
    {\normalfont \small Recall by 2 readers}   &{\normalfont \small 0.824} &{\normalfont \small 0.089} & {\normalfont \small 0.046}&  {\normalfont \small 0.041} \\ \hline 
    {\normalfont \small Recall by arbitration}   &{\normalfont \small 0.782} &{\normalfont \small 0.113} & {\normalfont \small 0.048}&  {\normalfont \small 0.056} \\ \hline 

\end{tcolorbox}
\end{table}

\subsection{Classifier performance}

During testing the same augmentations used during training were applied and the final predictions were averaged over 100 sample iterations.  Table~\ref{table-ranking} shows the sources of gain for MAMMO CNN and \textit{Classifier} diagnostic performance.  MV denotes the number of mammogram views used as input, which can be either a single view (1) or all views (4).  MTL is checked whenever multi-task learning was used.  If MTL is not checked, then the model was trained to only predict \textit{diagnosis} with no auxiliary prediction tasks.  TTA is checked whenever test-time augmentation was used (100 samples per patient). If TTA is not checked, then the AUROC and area under the precision-recall curve (AUPRC) were calculated over one sample prediction per patient. It is important to note that the reported AUROC values of 0.791 are relatively high compared to the existing state-of-the-art given the difficulty of the \textit{Tommy} dataset. For comparison we show our proposed method in comparison to the closest image-level CNN works of Zhang et. al \cite{fullimage-zhang} and Geras et al. \cite{krysztof-etal-2017} on the \textit{Tommy} dataset. We used the network and training methods provided in each respective publication.  We conducted additional experimentation of these methods and MAMMO CNN on the public CBIS-DDSM dataset, which is discussed further in Appendix~\ref{Appendix:Architecture}. For a better comparison, an AUPRC of 0.525 was reported for  \textit{Classifier}.  This is the first deep learning for breast cancer paper to report results in terms of AUPRC.  Because true negative counts are not a component in the calculation of precision and recall, AUPRC is a more appropriate metric than AUROC in situations of dataset imbalance with a high ratio of negative to positive samples and for evaluating screening, i.e., picking positives out of a population \cite{AUPRC-2015, AUPRC-2006}.
 
Fig.~\ref{fig:venn} shows the percentage of correct predictions between the radiologist versus \textit{Classifier} and is illustrated in more detail in Table~\ref{table-patientdistributions} over various sub-populations across the 1000 patient test set.  For example, consider the first row of all 1000 patients, the radiologists and \textit{Classifier} were both correct 82.9\% of the time as shown in the column $R^+C^+$ and were both wrong 3.5\% of the time as shown in column $R^-C^-$.
This table provides insight into when the radiologists are better, when \textit{Classifier} is better, and when they both agree on an outcome.  They have a high percentage of agreement on the common cases, such as patients with no sign of cancer, patients with no family history of breast cancer, and patients with non-suspicious mammograms.  Conversely, they have a low percentage of agreement on cases that are known to be challenging, such as patients with a family history of breast cancer, patients recalled by arbitration, patients with highly suspicious mammograms, and patients with mammograms containing spiculated masses, micro-calcifications, or distortions.

\begin{table*}[t!] 
\centering
\caption{ \label{table:final}Comparative performance of radiologist, $\textit{Classifier}$, and MAMMO on a test set of 1000 patients, TP is the true positive count, TN is the true negative count, FP is the false positive count and FN is the false negative count. MAMMO triage  performance gain is shown in comparison to the operating point \textit{Classifier}\textsuperscript{\textregistered}, which is a random allocation of patients (428) assigned to the \textit{Classifier} and the rest (572) to the radiologist.    Bold values denote the values that improve upon the radiologists performance alone. }
\begin{tcolorbox}[tab2,tabularx={p{2.6cm}|c|c|c|c|c|c|c|c}]
    {\normalfont \small \bf \textcolor{red!60!black}{Patient triage}} & 
    {\normalfont \small \bf \textcolor{red!60!black}{Radiologist patients}} &
    {\normalfont \small \bf \textcolor{red!60!black}{$\textit{Classifier}$ patients}} & 
    {\normalfont \small \bf \textcolor{red!60!black}{Cohen's $\kappa$}} &
    {\normalfont \small \bf \textcolor{red!60!black}{F1 score}} &
    {\normalfont \small \bf \textcolor{red!60!black}{TP}} & 
    {\normalfont \small \bf \textcolor{red!60!black}{TN}} &
    {\normalfont \small \bf \textcolor{red!60!black}{FP}} & 
    {\normalfont \small \bf \textcolor{red!60!black}{FN}}

    \\ \hline \hline
    {\normalfont \small Radiologist}   & {\normalfont \small 1000} & {\normalfont \small 0}& {\normalfont \small 0.708} & {\normalfont \small 0.755}&{\normalfont \small 120} & {\normalfont \small 802} & {\normalfont \small 42} & {\normalfont \small 36}  \\ \hline
    
    
    {\normalfont \small \textit{Classifier}}   & {\normalfont \small 0} & {\normalfont \small 1000}& {\normalfont \small 0.420} & {\normalfont \small 0.433} & {\normalfont \small 61} & {\normalfont \small \textbf{811}} & {\normalfont \small \textbf{33}} & {\normalfont \small 95} \\ \hline
    
    {\normalfont \small \textit{Classifier\textsuperscript{\textregistered}}}   & {\normalfont \small 572}& {\normalfont \small 428} & {\normalfont \small 0.570} & {\normalfont \small 0.633 } & {\normalfont \small 93} & {\normalfont \small 799} & {\normalfont \small 45} & {\normalfont \small 63} \\ \hline
     
     {\normalfont \small MAMMO }   & {\normalfont \small 572} & {\normalfont \small 428}& {\normalfont \small \textbf{0.716}} & {\normalfont \small \textbf{0.757}} & {\normalfont \small 120} & {\normalfont \small \textbf{803}} & {\normalfont \small \textbf{41}} & {\normalfont \small 36}  \\ \hline

    
\end{tcolorbox}
\end{table*}

\begin{figure*}[!htbp]
\centering
  \includegraphics[width=\linewidth]{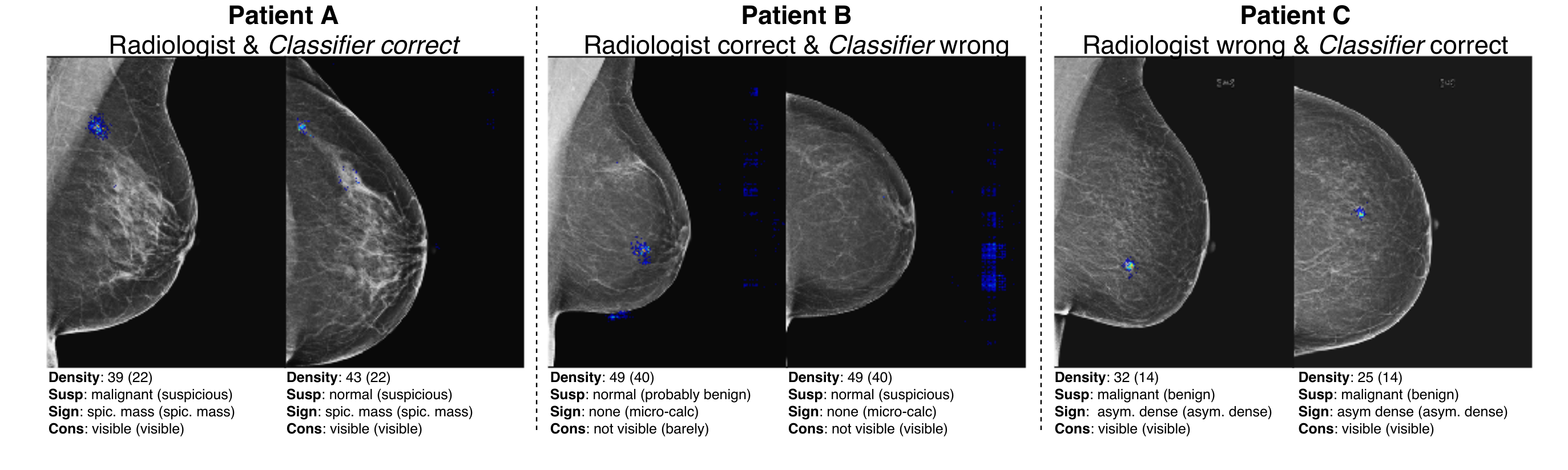}
  \caption{Example visualization of \textit{Classifier} on three positive (malignant) patients.  Patient A was diagnosed correctly by both the radiologists and \textit{Classifier}; Patient B was diagnosed correctly by the radiologists but not \textit{Classifier}; and Patient C was diagnosed correctly by \textit{Classifier}, but not the radiologists. For each patient, the malignant breast is shown with MLO view on the left and CC on the right.  The predicted \textit{density}, \textit{suspicion}, \textit{sign} and \textit{conspicuity} are shown.  The actual radiological annotations are in parenthesis. }
  \label{fig:viz}
\end{figure*}

\subsection{Triage performance}

Table~\ref{table:final} illustrates the comparative performance of the radiologists, \textit{Classifier}, and MAMMO on a 1000 patient subset of \textit{Tommy}.  We show performance gain of MAMMO in comparison to the radiologist using Cohen's kappa coefficient and the F1 score.
Cohen's kappa coefficient ($\kappa$) is reported and is a more reliable metric than percentage agreement because it takes into account the expected accuracy of random chance agreement \cite{cohen-kappa}.  
The F1 score, which is the harmonic mean of precision and recall, does not factor in true negatives into it's calculation and is therefore a more appropriate and reliable metric in breast cancer screening for similar reasons presented earlier regarding AUPRC.  
Table~\ref{table:final} shows four scenarios over the 1000 patient \textit{Tommy} test set.
The first scenario is a single-reading scenario where the radiologist reads all 1000 patients.  The second scenario is \textit{Classifier} in a single-reading scenario where all the patients are read by the \textit{Classifier}.  This is the most common application in the existing literature in machine learning for breast cancer. The third scenario, labeled  \textit{Classifier\textsuperscript{\textregistered}}, is a random allocation of patients (572) read by the radiologist and the rest (428) by the \textit{Classifier}. The last scenario shows MAMMO minimizing the number of patients read by the radiologist. In this scenario, \textit{Classifier} reads and filters 428 patients from the radiologist with significantly better performance than the random triage of  (\textit{Classifier\textsuperscript{\textregistered}}) and does not degrade the collective performance of \textit{Classifier} and the radiologist in regards to any of the presented metrics. Additional details and operating points are covered in Appendix~\ref{Appendix:triage}.

\subsection{Visualizing MAMMO}

Visualizing the processing of a CNN is critical for understanding and interpreting model effectiveness and fidelity.  Although several methods exist for visualization in CNNs \cite{visualizing1, visualizing2, visualizing3}, most require large data sets and network retraining.  Instead, the method proposed in \cite{krysztof-etal-2017} was used, which did not require network retraining and worked by simply examining the network's output sensitivity to perturbations in each input pixel.  The premise was that a higher output variance will be observed when an ``important'' input pixel is perturbed.  Using this method, Fig.~\ref{fig:viz} shows an example on three positive patients.   Patient A was diagnosed correctly by both the radiologist and \textit{Classifier}.  For this patient, all of the predictions by \textit{Classifier} agreed with the outcome except for the suspicion of the CC view, which \textit{Classifier} deemed normal instead of suspicious. Patient B was diagnosed correctly by the radiologist but not \textit{Classifier}. Patient C was diagnosed correctly by \textit{Classifier} but not the radiologist.  For this patient, the malignant lesion correctly identified by \textit{Classifier} was also discovered by the radiologist, but was misdiagnosed as benign.  The visualization for Patient B is the only patient with background pixels highlighted which agrees with the negative (normal) predictions of \textit{Classifier}.  For Patients A and C, \textit{Classifier} recognized at least 1 ``well-defined'' region in either view and did not have any visible background pixels highlighted.  

\begin{table*}[htbp] 
\centering
\caption{ \label{table:workload} Further stratification of patient populations to illustrate workload of MAMMO versus radiologist.  Displayed are the percentage of total patients read by the radiologist with respective kappa coefficients and F1 scores.  }
\begin{tcolorbox}[tab2,tabularx={p{4.2cm}|c|c|c|c|c|c}]
    {\normalfont \small \bf \textcolor{red!60!black}{Population}} & 
    {\normalfont \small \bf \textcolor{red!60!black}{Total patients}} & 
    {\normalfont \small \bf \textcolor{red!60!black}{\% to Rad.}} &
    {\normalfont \small \bf \textcolor{red!60!black}{Rad. $\kappa$ }} &
    {\normalfont \small \bf \textcolor{red!60!black}{MAMMO $\kappa$ }} &
    {\normalfont \small \bf \textcolor{red!60!black}{Rad. F1}} &
    {\normalfont \small \bf \textcolor{red!60!black}{MAMMO F1}} 
\\ \hline \hline
    {\normalfont \small All patients}               & {\normalfont \small 1000} & {\normalfont \small 42.8} & {\normalfont \small 0.708} &{\normalfont \small 0.716} &{\normalfont \small 0.755} &  {\normalfont \small 0.757}   \\ \hline \hline
    {\normalfont \small High conspicuity}           &{\normalfont \small 441}   & {\normalfont \small 47.0} & {\normalfont \small 0.702}  & {\normalfont \small 0.701} & {\normalfont \small 0.783}& {\normalfont \small 0.763}  \\ \hline 
    {\normalfont \small Low conspicuity}            &{\normalfont \small 559}   & {\normalfont \small 37.2} & {\normalfont \small 0.641} & {\normalfont \small 0.728}  & {\normalfont \small 0.667}& {\normalfont \small 0.739} \\ \hline \hline
    {\normalfont \small No sign of cancer}          &{\normalfont \small 314}   & {\normalfont \small 31.2} & {\normalfont \small 68.8}  & {\normalfont \small 0.987} & {\normalfont \small 1.000}& {\normalfont \small 0.981}  \\ \hline 
    {\normalfont \small Circumscribed mass}         &{\normalfont \small 226}   & {\normalfont \small 42.4} & {\normalfont \small 0.535} & {\normalfont \small 0.624} & {\normalfont \small 0.560}& {\normalfont \small 0.667}  \\ \hline 
    {\normalfont \small Spiculated mass}            &{\normalfont \small 78}    & {\normalfont \small 55.6} & {\normalfont \small 0.291}& {\normalfont \small 0.265} & {\normalfont \small 0.924}& {\normalfont \small 0.900}  \\ \hline  
    {\normalfont \small Micro-calcification}        &{\normalfont \small 153}   & {\normalfont \small 42.3} & {\normalfont \small 0.433} & {\normalfont \small 0.495} & {\normalfont \small 0.567}& {\normalfont \small 0.590}  \\ \hline  
    {\normalfont \small Distortion}                 &{\normalfont \small 42}    & {\normalfont \small 59.5} & {\normalfont \small 0.583}& {\normalfont \small 0.716} & {\normalfont \small 0.791}& {\normalfont \small 0.850}  \\ \hline 
    {\normalfont \small Asymmetrical density}       &{\normalfont \small 187 }  & {\normalfont \small 42.1} & {\normalfont \small 0.640} & {\normalfont \small 0.732} & {\normalfont \small 0.682}& {\normalfont \small 0.762}  \\ \hline \hline 
    {\normalfont \small High breast density}        &{\normalfont \small 491}   & {\normalfont \small 43.8} & {\normalfont \small 0.710} &{\normalfont \small 0.703} &{\normalfont \small 0.755}&  {\normalfont \small 0.753}  \\ \hline 
    {\normalfont \small Low breast density}         &{\normalfont \small 509}   & {\normalfont \small 39.3} & {\normalfont \small 0.706} &{\normalfont \small 0.740}  &{\normalfont \small 0.754} &   {\normalfont \small 0.760}    \\ \hline  \hline
    {\normalfont \small Family History}             &{\normalfont \small 164}   & {\normalfont \small 46.4} & {\normalfont \small 0.461}      &{\normalfont \small 0.495} &{\normalfont \small 0.841}& {\normalfont \small 0.812} \\ \hline
    {\normalfont \small No Family History}          &{\normalfont \small 836}   & {\normalfont \small 40.6} & {\normalfont \small 0.494}    &{\normalfont \small 0.606} &{\normalfont \small 0.517}& {\normalfont \small 0.603} \\ \hline\hline 
    {\normalfont \small Age over $\mu$ ($\geq$ 57)} &{\normalfont \small 441}   & {\normalfont \small 46.7} & {\normalfont \small 0.722} &{\normalfont \small 0.706} &{\normalfont \small 0.789}& {\normalfont \small 0.768} \\ \hline
    {\normalfont \small Age under $\mu$ (< 57)}     &{\normalfont \small 559}   & {\normalfont \small 37.4} & {\normalfont \small 0.654} & {\normalfont \small 0.720} & {\normalfont \small 0.686}& {\normalfont \small 0.733}  \\ \hline \hline
    {\normalfont \small Recall by 1 reader}         &{\normalfont \small 261}   & {\normalfont \small 39.5} & {\normalfont \small 0.774}& {\normalfont \small 0.714} & {\normalfont \small 0.800}& {\normalfont \small 0.750}  \\ \hline
    {\normalfont \small Recall by 2 readers}        &{\normalfont \small 615}   & {\normalfont \small 41.6} & {\normalfont \small 0.690} & {\normalfont \small 0.708} & {\normalfont \small 0.741}& {\normalfont \small 0.747}  \\ \hline 
    {\normalfont \small Recall by arbitration}      &{\normalfont \small 124}   & {\normalfont \small 45.2} & {\normalfont \small 0.688} & {\normalfont \small 0.746} & {\normalfont \small 0.755}& {\normalfont \small 0.801}  
\end{tcolorbox}
\end{table*}
        
\begin{table}[!htbp]
\centering
\caption{ \label{table:age} Average of the variance between all mammogram views for \textit{Classifier} predicted breast \textit{density} and predicted patient \textit{age} reported as the mean breast density variance (MBDV) and mean age variance (MAV).}
\begin{tcolorbox}[tab2,tabularx={p{3.4cm}|c|c}]
    {\normalfont \small \bf \textcolor{red!60!black}{Population}} & 
    {\normalfont \small \bf  \textcolor{red!60!black}{MBDV (\%)}} &
    {\normalfont \small \bf \textcolor{red!60!black}{MAV (years)}} 
    \\ \hline \hline
    {\normalfont \small All patients}   & {\normalfont \small 4.614} & {\normalfont \small  4.365} \\ \hline \hline
    
    
    
    {\normalfont \small \textit{Classifier} positive}   & {\normalfont \small 18.492} & {\normalfont \small 10.448} \\ \hline 
    {\normalfont \small \textit{Classifier} negative}   & {\normalfont \small 3.505} & {\normalfont \small 3.880} \\ \hline

\end{tcolorbox}
\end{table}

\subsection{Analyzing MAMMO} 
MAMMO opens the door for many important research questions and potential discoveries from both a machine learning and medical perspective.  Consider Table~\ref{table:workload} that illustrates the patient distribution between MAMMO \textit{Classifier} and the radiologist over various sub-populations.  From this table, MAMMO filters or screens the simpler cases, i.e., patients with lower breast density, no family history, no sign of cancer, etc., from the radiologist.  \textit{Classifier} screened the highest percentage ($>60\%$) of patients for the following populations: patients with low breast density, younger patients, patients recalled by one reader,  patients with low conspicuity, and patients with no sign of cancer in their mammograms. Conversely, MAMMO defers the more complex cases over to the radiologists. The \textit{Classifier} screened a lower percentage of patients for the following populations: patients with high breast density, older patients, patients recalled by arbitration between 2 readers, and patients exhibiting spiculated masses, distortions, high conspicuity, or high suspicions in their mammograms.  Lastly, Table~\ref{table:final} and ~\ref{table:workload} demonstrates the performance improvements in $\kappa$ and F1 score that reflect the overall improvement of MAMMO over the radiologists across the entire population and a majority of the sub-populations. 

Past studies demonstrated that asymmetry (in terms of density) between breasts are often indicative of cancer \cite{breast-asymmetry1, breast-asymmetry2}. Table~\ref{table:age} shows the average variance between the predicted \textit{density} and \textit{age} for each mammogram view.  For example, this table shows that for patients predicted positive (malignant) by \textit{Classifier} the average variance of the predicted \textit{density} over all four mammogram views (MLO right, MLO left, CC right and CC  left) is much higher at  18.49\% compared to 3.505\% for negative patients. The data provides valuable insight into the validity of \textit{Classifier}'s predictions in relation to medical findings and demonstrates the diagnostic advantages multiple mammogram views provides over a single mammogram. \textit{Age} was also included in this table because of it's known association with breast density and malignancy \cite{age-breastdensity, age-breastdensity2}.


\begin{figure*}[!t]
  \includegraphics[width=\linewidth]{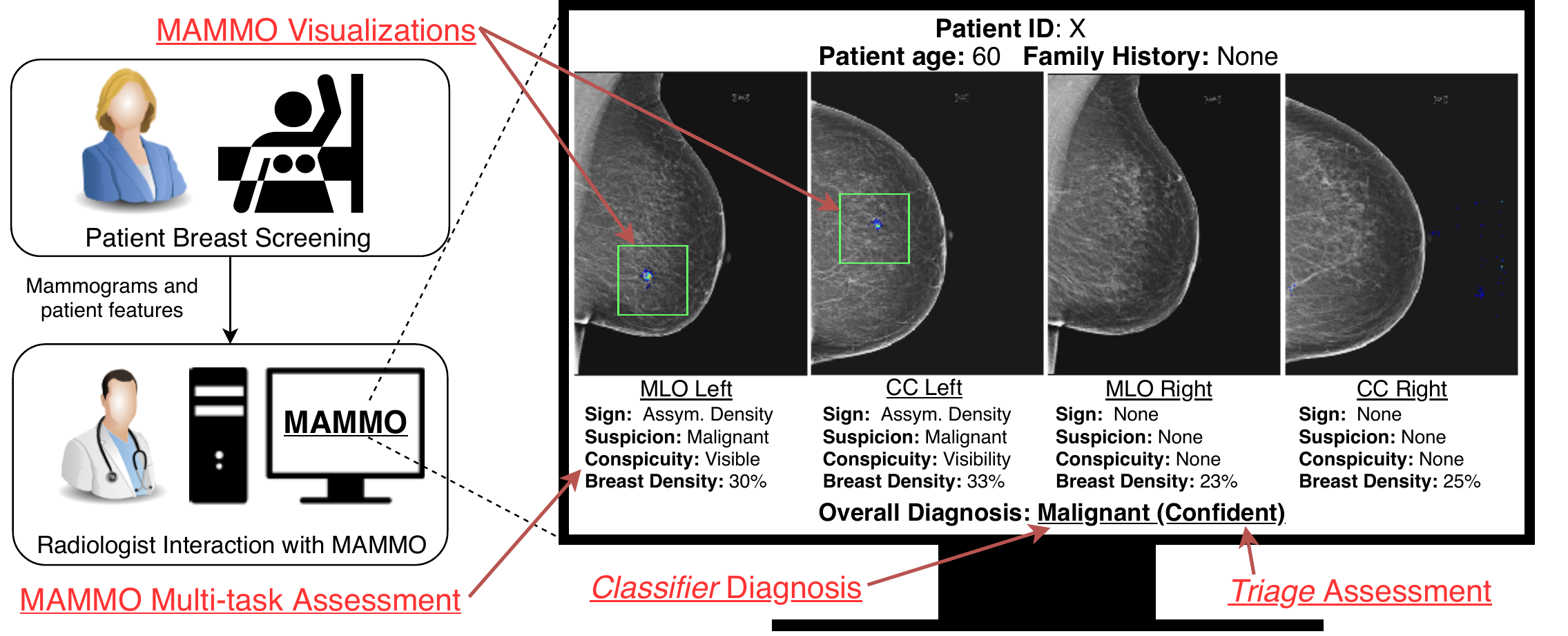}
  \caption{Example illustration of MAMMO interpretability by the radiologist in two ways: 1) visualization of MAMMO CNN identified features, and 2) multi-task outputs provide additional assessments for radiologist to scrutinize.}
  \label{fig:dashboard}
\end{figure*}

\section{MAMMO In Practice}

In our collaboration with radiologists, we have identified an AI system’s ability to assist and parallel the decisions of radiologists as key requirements for its acceptance in clinical practice.  This is why we have designed our approach to issue not only cancer predictions, but also radiological assessments such as the conspicuity, suspicion, breast density, etc., in a similar manner as a radiologist would make an assessment. This allows our approach to provide radiologists more interpretable predictions and estimates, thereby enabling better human-machine collaboration for mammography.  MAMMO provides interpretability not currently offered in the machine learning for breast cancer literature.  Fig.~\ref{fig:dashboard} shows an example of how MAMMO provides additional information that can be used to debug and interpret MAMMO (both \textit{Classifier} and \textit{Triage}) decisions.  Existing methods in machine learning for mammography provide visualizations, but do not have the ability to provide the multi-task annotations that MAMMO is capable of.  The multi-task outputs are extracted imaging features that emulate radiological assessment and are what a radiologist would naturally consider when examining multiple mammogram views.  For example, breast density asymmetries between left and right breast are often indicators of cancer \cite{breast-asymmetry1}.

\section{Conclusion}

Our approach addresses a novel problem, i.e., automatically and confidently triaging mammograms into ones which need to be read by a radiologist and ones that do not, thereby saving scarce clinical resources (radiologists’ time). We presented a first approach for such a triaging system, which has the potential to save countless hours for overworked radiologists and provide them more time to focus on the difficult and complex cases that warrant additional scrutiny. In addition, we are the first to introduce MTL in image-level mammogram classification with two objectives: 1) improving the predictive performance of CNN approaches, and 2) improving the interpretability and usability of AI approaches by predicting both malignancy and radiological labels known to be associated with cancer (e.g. conspicuity, suspicion etc.) that a radiologist could debug and question to better interact with and interpret MAMMO issued predictions.  Finally, we tested our models on one of the most difficult datasets in the current literature, and observed that MAMMO filtered the patient sub-populations that are associated with lower-risk from the radiologists.   
The approach used in MAMMO pushes the frontier of human and artificial intelligence synergism and is applicable to many other medical imaging modalities, including MRI, CT, pathology, etc., and to countless applications extending beyond the health-care domain.

\appendices

\section{\label{Appendix:related}Related works}

There are two primary objectives in the existing literature for machine learning and mammography.  The first is to assist the radiologists reading through CAD techniques.  The objective here is to assist the radiologist in making decisions, but does not allow for any patient to bypass radiologist reading.  The second objective, which has recently gained popularity, is to train CNNs to diagnose a patient without radiologist reading.  Both are addressed in the following subsections.  

\subsection{CAD}

CAD was introduced into screening mammography nearly 30 years ago.  These traditional approaches relied on conventional techniques centered around hand-crafted imaging features \cite{jamieson-2012}.  The majority of existing literature of CNNs in CAD are used for density classification \cite{fonseca-2015, kallenberg-2016, wu-etal-2017, ahn-2017, mohamed-2018}, segmentation \cite{ dhungel-2015c, zhu-2016, demoor-2018}, or improving detection rates of malignant soft tissues or micro-calcifications. \cite{kooi-2016} used a random forest classifier over outputs of a CNN for mass detection for CAD.  They demonstrated the importance of class balancing equal numbers of positive and negative samples, which was applied in this work.   \cite{abbas-2016, jiao-2016} presented CNNs for tumor or lesion classification for CAD in mammography and \cite{samala-2016b, bekker-2016} presented CNNs for micro-calcification classification for CAD in mammography.  \cite{suzuki-2016, huynh-2016} first used transfer learning for training CNNs for CAD in mammography. \cite{ribli-2017} uses Faster R-CNN for CAD in mammography and reports the highest  AUROC of 0.95 on the INbreast dataset.  
Though these works improved on traditional methods based on hand-crafted features, they still do not allow patients to bypass radiologist reading.  Although initial studies on CAD efficacy showed improved performance for breast cancer diagnosis \cite{dheeba-2014}, these results were later disputed by larger and more comprehensive studies.  The largest study conducted reported that CAD did not statistically improve diagnostic performance of mammograms across any population because radiologists either relied only on CAD or ignored it \cite{lehman-2015}. Additionally, \cite{nishikawa-2018} reported two flaws with current CAD in mammography: 1) radiologist often ignore the majority (71 percent) of correct computer detection of a cancer on a mammogram and 2) radiologists are not using screening mammography CAD as intended, which is as a second reader.   A recent publication by \cite{karssemeijer-deeplearning} showed radiologists improved performance improvements when using a deep learning computer system as decision support in breast cancer diagnosis.  With this in mind, MAMMO was designed to function as a pre-screening filter for the radiologists, rather than a tool (like CAD) that could be overlooked and misused.

\begin{table*}[!htbp] 
\centering
\caption{ \label{table:comparisons}Comparison of related image-level and multi-view architectures.  Bold represents the proposed method. * denotes $\mu$ AUROC. MGV represents the number of mammogram views used as input. ROI is Y if either ROI or segmentation masks were needed for training or inference, otherwise N.  MTL is Y if multitask learning used, otherwise N.  MAMMO \textit{Classifier} is shown in bold.}
\begin{tcolorbox}[tab2,tabularx={p{3.9cm}|c|c|c|c|c|c|c}]
    {\normalfont \small \bf \textcolor{red!60!black}{Method}} & 
    {\normalfont \small \bf \textcolor{red!60!black}{Dataset}} &
    {\normalfont \small \bf \textcolor{red!60!black}{Num. Patients}} & 
    {\normalfont \small \bf \textcolor{red!60!black}{Task}} & 
    {\normalfont \small \bf \textcolor{red!60!black}{MGV}} & 
    {\normalfont \small \bf \textcolor{red!60!black}{ROI}} & 
    {\normalfont \small \bf \textcolor{red!60!black}{MTL}} & 
    {\normalfont \small \bf \textcolor{red!60!black}{AUROC}} 
    \\ \hline \hline
    {\normalfont \small Geras et al. \cite{krysztof-etal-2017}}   & {\normalfont \small private} & {\normalfont \small 201698}& {\normalfont \small BI-RADS} & {\normalfont \small 4}& {\normalfont \small N}&{\normalfont \small N}&{\normalfont \small 0.678*}\\ \hline
     {\normalfont \small Akselrod-balin et al. \cite{akselrod-2016}}   & {\normalfont \small private} & {\normalfont \small 300}& {\normalfont \small BI-RADS} & {\normalfont \small 1}& {\normalfont \small N}&{\normalfont \small N}&{\normalfont \small 0.78 acc.} \\ \hline
    {\normalfont \small Carneiro et al. \cite{carneiro-2015}}   & {\normalfont \small DDSM/INbreast} & {\normalfont \small 287}& {\normalfont \small BI-RADS} & {\normalfont \small 2}& {\normalfont \small Y}&{\normalfont \small N}&{\normalfont \small 0.91}  \\ \hline
    {\normalfont \small Carneiro et al. \cite{carneiro-2017}}   & {\normalfont \small INbreast} & {\normalfont \small 115}& {\normalfont \small malignancy} & {\normalfont \small2 }& {\normalfont \small Y}&{\normalfont \small N}&{\normalfont \small 0.860} \\ \hline
    {\normalfont \small Bekker et al. \cite{bekker-2016}}   & {\normalfont \small DDSM} & {\normalfont \small 172}& {\normalfont \small malignancy} & {\normalfont \small 2}&{\normalfont \small Y}&{\normalfont \small N}&{\normalfont \small 0.800} \\ \hline
    
    {\normalfont \small Dhungel et al. \cite{dhungel-2017}}   & {\normalfont \small INbreast} & {\normalfont \small 115}& {\normalfont \small malignancy} & {\normalfont \small 2}& {\normalfont \small Y}&{\normalfont \small N}&{\normalfont \small 0.800} \\ \hline
    {\normalfont \small \textbf{MAMMO $\textit{Classifier}$}}   & {\normalfont \small \textbf{Tommy}} & {\normalfont \small \textbf{8162}}& {\normalfont \small \textbf{malignancy}} & {\normalfont \small \textbf{\boldmath{4}}}& {\normalfont \small \textbf{N}} &{\normalfont \small \textbf{Y}} &{\normalfont \small \textbf{0.791}}\\
\end{tcolorbox}
\end{table*}
\subsection{Radiologist-machine collaboration}

The success of CNNs across the computer vision domain has lead to many applications and significant improvements in the medical imaging field.  In mammography, these works are centered around detection, classification, or both.  The most popular task in this domain is to diagnose cancer or predict BI-RADS and compare performance against radiologist.  A majority of the existing literature where CNNs are applied in this field is grouped into two categories: 1) using ROI-based networks, and 2) image-level methods that do not rely on ROI. 

\subsubsection{ROI-based methods}
Due to the large amount of data required for training large CNNs and the limited number of available datasets, many implementations require utilization of ROIs or segmentation masks for maximizing performance. State of the art implementations, utilizing region proposal networks, sliding windows, or patch classifiers for mammogram diagnosis rely on radiologist labeled ROIs and often have the highest reported diagnostic performance \cite{shen-2017, jiao-2016, akselrod-2016, becker-2016, platania-2017, teare-2017, kooi-2017,akselrod-2017,carneiro-2017,jadoon-2017, hepsag-2017, dream-2017, ribli-2017, mohamed-2018, multi-task2}.  However, all of these works rely on a very scarce and costly commodity, i.e., a dataset with cancer locations identified. ROI-based approaches have disadvantages other than the limitation of available location-annotated datasets.  First, high-level contextual features external to the ROI are not learned \cite{krysztof-etal-2017}.  Secondly, in high noise scenarios where breast density may hide a visible tumor, a radiologist considers macroscopic features, such as asymmetry between breasts or subtle changes in mammograms from previous examination, to assist in malignancy diagnosis \cite{breast-asymmetry2, breast-asymmetry1}.  

\subsubsection{Image-level methods}

Networks that are trained with full images have been shown to improve diagnostic performance, but require more training data \cite{end2end-2017}.  In comparison to ROI-based methods, the training data does not require annotated locations that makes data acquisition a lot simpler, cheaper and scalable.  The work of \cite{krysztof-etal-2017} presented the richest mammography datasets used (with over 200,000 mammograms).  Because of this, they are one of the few researchers who attempt an image-level approach utilizing all four mammogram views to predict BI-RADS score.  Our dataset is \textit{significantly} smaller, but we draw motivation from their work of using all four mammogram views without relying on any ROI.  Results are compared to theirs for a benchmark comparison.  Other related image-level and multi-view networks were presented in \cite{akselrod-2016, carneiro-2015, carneiro-2017, zhu-2016, bekker-2016} and are shown in Table~\ref{table:comparisons} for comparison.  

MTL has been successfully used on an ROI level in mammography \cite{kisilev-2016, multi-task2}, but this work is the first to apply MTL to image-level mammogram classification.  \cite{carneiro-2015, bekker-2016, krysztof-etal-2017, yi-2017}  used multiple views for improving classification performance, however this work is the first to do so by concatenating the multi-task outputs of each mammogram view.  Many early investigative works have shown the success of transfer learning using non-medical or natural images to classify mammograms  \cite{argyriou-2006}. Specifically, these publications have shown performance gains from using models pre-trained with ImageNet weights, such as AlexNet, Inception or ResNet\cite{huynh-2016,levy-2016, samala-2016a, jiang-2017, yi-2017}. Motivated by their success, a preliminary trade study was conducted between ResNet50, VGG16, VGG19, InceptionV3, InceptionResNetV2, and Xception to select the highest performing model \cite{resnet-2015, vgg-2014, inceptionv3-2015, inceptionresnet-2016, xception-2016}. Additional details and results are reported in Appendix A.  

The closest related works are presented in Table~\ref{table:comparisons}, and although it is difficult to draw a direct comparison to these works, we highlight the limitations of existing works in comparison to ours. MAMMO \textit{Classifier} has a reported AUROC of 0.791 and is significantly higher than \cite{krysztof-etal-2017} at 0.678, who predicted BI-RADS (0, 1, and 2).  The works of \cite{carneiro-2017}, \cite{dhungel-2015c}, and \cite{zhu-2016} use the INbreast dataset to predict malignancy and have marginally higher AUROC than we do at 0.8 to 0.86.  However, because of their small dataset size of 115 patients, the reported results could be subject to high variance.  Additionally, this dataset does not have the challenging overlapping tissues present in the \textit{Tommy} dataset.  

We believe this to be the first work in deep learning for mammography to report results in terms of AUPRC, which has several unique advantages over AUROC.  Many real-world examples contain a lot more negative cases than positive.  In these situations of large class imbalance, AUPRC is favored since the true negative count is not factored in it's calculation \cite{AUPRC-2006}.  This makes AUPRC a better metric for comparison between two different datasets that do not contain the same balance.  Additionally, AUROC does not take into account prevalence, i.e., when prevalence is very low, even a ``high'' AUROC may result in low post-test probability \cite{AUPRC-2015}.

\begin{table*}[!htbp] 
\centering
\caption{ \label{table-summary} A trade-study of candidate CNN architectures on public CBIS-DDSM dataset to select the best CNN from available pre-trained ImageNet models to use as MAMMO CNN.  For the ROI trade study, the reported values are the AUROC for 2-class, 3-class and 5-class stratifications.  For full-images the AUROC and AUPRC are reported.  The highest values for each experiment are in bold.}
\begin{tcolorbox}[tab2,tabularx={p{3.1cm}||c|c|c||c|c}]
    {\normalfont \small \bf \textcolor{red!60!black}{Model}} & 
    {\normalfont \small \bf  \textcolor{red!60!black}{ROI 2-class}} &
    {\normalfont \small \bf \textcolor{red!60!black}{ROI 3-class}} &
    {\normalfont \small \bf \textcolor{red!60!black}{ROI 5-class}} & 
    
    {\normalfont \small \bf \textcolor{red!60!black}{Full-image AUROC}} &
    {\normalfont \small \bf \textcolor{red!60!black}{Full-image AUPRC}}
    \\ \hline \hline
    {\normalfont \small ResNet50}   & {\normalfont \small 0.740} & {\normalfont \small 0.734} & {\normalfont \small 0.706} & {\normalfont \small 0.607} & {\normalfont \small 0.488}\\ \hline
    {\normalfont \small VGG16}   & {\normalfont \small 0.762} & {\normalfont \small 0.741} & {\normalfont \small 0.679} & {\normalfont \small 0.538} & {\normalfont \small 0.432}\\ \hline
    {\normalfont \small VGG19}   & {\normalfont \small 0.783} & {\normalfont \small 0.739} & {\normalfont \small 0.665} & {\normalfont \small 0.542} & {\normalfont \small 0.402}\\ \hline
    {\normalfont \small InceptionV3}   & {\normalfont \small 0.800} & {\normalfont \small 0.731} & {\normalfont \small 0.712} & {\normalfont \small 0.640} & {\normalfont \small \textbf{0.541}}\\ \hline
    {\normalfont \small InceptionResNetV2}   & {\normalfont \small \textbf{0.842}} & {\normalfont \small \textbf{0.841}} & {\normalfont \small \textbf{0.844}} & {\normalfont \small \textbf{0.652}} & {\normalfont \small 0.493}\\ \hline
    {\normalfont \small Xception}   & {\normalfont \small 0.767} & {\normalfont \small 0.706} & {\normalfont \small 0.741} & {\normalfont \small 0.565} & {\normalfont \small 0.434}\\ \hline

\end{tcolorbox}
\end{table*}

\begin{table}[!htbp] 
\centering
\caption{ \label{table-ddsm}Comparison of models on public dataset DDSM.  For consistency all models are shown using TTA.  Each model is trained using their published hyperparameters.}
\begin{tcolorbox}[tab2,tabularx={p{4.4cm}|c|c}]
    {\normalfont \small \bf \textcolor{red!60!black}{Model}} & 
    {\normalfont \small \bf \textcolor{red!60!black}{AUROC}} &
    {\normalfont \small \bf \textcolor{red!60!black}{AUPRC}}
    \\ \hline \hline
    {\normalfont \small Geras et. al \cite{krysztof-etal-2017}}   & {\normalfont \small 0.490} & {\normalfont \small 0.408} \\ \hline
    {\normalfont \small Zhang et. al \cite{fullimage-zhang}}   & {\normalfont \small 0.531} & {\normalfont \small 0.423} \\ \hline
    {\normalfont \small MAMMO CNN}   & {\normalfont \small 0.652} & {\normalfont \small 0.493} \\ \hline
\end{tcolorbox}
\end{table}

\section{\label{Appendix:candidate} Candidate CNN selection}

Transfer learning utilizing pretrained models on non-medical datasets has been shown to have competitive, and sometimes state-of-the-art, performance in many medical imaging and mammography tasks \cite{khosravi-2018, habibzadeh-2018, jiao-2016, carneiro-2017, carneiro-2015, huynh-2016, levy-2016}. Recent deep learning toolkits, such as \textit{Keras}, allow practitioners to fine-tune and utilize many of the successful ImageNet models with ease \cite{keras-2015}.  Because of this, we chose to evaluate and select the best CNN from the following ImageNet algorithms: ResNet50, VGG16, VGG19, InceptionV3, InceptionResNetV2 and Xception. 
We judged model performance on both ROI and full-images from the Curated Breast Imaging Subset of the Digital Database of Screening Mammography (CBIS-DDSM) \cite{ddsm-database1, ddsm-database2}.  The DDSM is a database of 2,620 scanned film mammography studies. It contains normal, benign, and malignant cases with verified pathology information. The CBIS-DDSM collection includes a subset of the DDSM data selected and curated by a trained mammography reader.  We chose this database due to the large number of related works using it, particularly with ROIs \cite{becker-2016, carneiro-2015, carneiro-2017, shen-2017, levy-2016, dhungel-2015a, jiao-2016, zhu-2016, dream-2017}. 

We emulated the methodology presented in \cite{shen-2017}, a finalist in the 2016 DREAM mammography challenge, who generated a full-field mammogram classifier by first pre-training on ROIs from CBIS-DDSM.  
In the first step, we extracted patches from full-field mammograms without down-scaling and saved the images as 224 x 224 8-bit PNG files. 
Before saving patches we also standardized (0 $\mu$, 1 $\sigma$) the entire set of patches by performing pixel-wise subtraction of the dataset mean and dividing by the dataset standard deviation.  
For every ROI patch saved we generated a ``background'' image, which was a uniformly random sampled region on the opposite (vertical and horizontal) half of the image.  For training and testing we used an approximate 90-10 split, where 4000 total patches (including backgrounds) were used to train our network.  
We used an approximate 1:1 ratio for masses/calcification to background images.  To deal with an extremely small training-set size and mitigating over-fitting, we applied random augmentation to each training image with the following specification: rotation within $\pm$25 degrees, shear up to 20 degrees counter-clockwise,  horizontal flips, vertical flips, and zoom within $\pm$10\%.  We used a batch size of 16 and a cross entropy loss function. 
An iterative multi-step approach was used in training each CNN.  The \textit{Adam} optimizer with a learning rate of $10^{-3}$ was used for training the top layer, a learning rate of $10^{-4}$ for the top 50\% of the network, and a learning rate of $10^{-5}$ for fine tuning the rest of the network as described in \cite{yi-2017, levy-2016, shen-2017}.  For full-image experimentation, we used the same preprocessing, network hyperparameters and architecture used for ROIs, except we did not randomly sample background patches and also resized mammograms to 320 x 416 to preserve the aspect ratio.  

Table~\ref{table-summary} shows a comparison of candidate CNN architectures used to evaluate and test our approach.  We evaluated 3 different class partitions for ROI images.  In the 2-class experiment, ROI were classified as either benign or malignant. In the 3-class experiment, ROI were classified as either background, benign or malignant.  And in the 5-class experiment, ROI were classified as one of background, benign calcification, benign mass, malignant calcification or malignant mass.  Each of the neural networks were initialized with pre-trained ImageNet weights, and had the top-layer replaced by a global average pooling layer followed by a new fully-connected dense classifier. A single dense layer of 1024 neurons was selected to bias model fitting into the convolutional layers. Hyper-parameter tuning was forgone, since the goal of this experiment was just a ranking system for CNN selection. InceptionResNetV2 performed the best in each classification task and metric, other than image-level AUPRC.

Table~\ref{table-ddsm} shows the single mammogram classification performance of MAMMO CNN to the closely related works of Zhang et. al \cite{fullimage-zhang} and Geras et al. \cite{krysztof-etal-2017} on the public DDSM dataset.  Each model used the same image preprocessing and augmentation presented in this section, and was trained using their published training hyperparameters and architecture.  Slight modification of the network used in \cite{krysztof-etal-2017} was required to accommodate a single mammogram rather than all four mammogram views.  This was done by simply providing all mammograms into the first CNN they used and keeping the same subsequent layers unmodified.

\section{\label{Appendix:Architecture} Details of Deep Learning Implementation}
\subsection{CNN architecture}
Much of this work was motivated by the multi-view CNN presented in \cite{krysztof-etal-2017}. For the purpose of comparing experimental results, the same final non-convolutional layers were used.  Consider the MAMMO CNN shown in Fig.~\ref{fig:mammocnn}, the top dense layer was removed and replaced with a global average pooling (GAP) layer, allowing for input dimension variation, followed by a dropout layer with a drop-rate of 0.2 and a dense layer of 1024 neurons with a rectified linear unit (ReLU) activation function. The total number of output neurons for each MAMMO CNN was 19, which correspond directly to the number of labels presented in Table~\ref{table-tommy}.

The outputs of four MAMMO CNN networks were concatenated to generate  \textit{Classifier} and \textit{Triage}.  Each mammogram view was passed into a designated input view or channel as shown in, for example, all MLO right mammograms are passed into the first input mammogram slot.  For both \textit{Classifier} and \textit{Triage}, MAMMO CNNs were concatenated using a standard concatenate layer followed by 4 dense layers of 2048, 1024, 512, and 256 neurons, each utilizing ReLU activation and glorot uniform initialization.  In between each dense layer a dropout at a rate of 0.2 was applied.  The final top-most layer was a 2-neuron dense layer for both and was initialized with glorot uniform weighting and soft-max activation.

\begin{table}[t!]
\centering
\caption{ \label{table-tommy} Multi-task target output descriptions. * denotes the task was a regression, rather than a categorical task.}
\begin{tcolorbox}[tab2,tabularx={l|l|l}]
    {\normalfont \small \bf \textcolor{red!60!black}{Task}} & 
    {\normalfont \small \bf \textcolor{red!60!black}{Source}} &
    {\normalfont \small \bf  \textcolor{red!60!black}{Label/Description}} 

    \\ \hline \hline
    {\normalfont \small Diagnosis} & {\normalfont \small biopsy}  & {\normalfont \small malignant or benign} 
    \\ \hline 
    {\normalfont \small Sign} & {\normalfont \small radiologist}  & {\normalfont \small none, circumscribed,}  \\
    {\normalfont \small }  & {\normalfont \small} & {\normalfont \small spiculated, calcification,}  \\
    {\normalfont \small}   & {\normalfont \small }& {\normalfont \small distortion, or asymmetrical} 
    \\ \hline
    {\normalfont \small Suspicion}  & {\normalfont \small radiologist} & {\normalfont \small normal, benign, probably} \\
      {\normalfont \small }   & {\normalfont \small }& {\normalfont \small benign, suspicious, or} \\
    {\normalfont \small }  & {\normalfont \small } & {\normalfont \small malignant} \\ \hline

    {\normalfont \small Conspicuity}  & {\normalfont \small radiologist} & {\normalfont \small not visible, barely visible,}  \\
    {\normalfont \small }   & {\normalfont \small } & {\normalfont \small not clearly visible, or}  \\
    {\normalfont \small }    & {\normalfont \small }& {\normalfont \small clearly visible,} \\\hline
    {\normalfont \small Breast density*}   & {\normalfont \small radiologist} & {\normalfont \small assessed along 10-cm VAS} \\ \hline 
    {\normalfont \small Age*}  & {\normalfont \small patient}  & {\normalfont \small age at mammogram} \\ \hline  

\end{tcolorbox}
\end{table}

\subsection{Training details}

To bias \textit{diagnosis} as the primary objective, loss weighting was adjusted according to the auxiliary output losses, such that the loss weight for \textit{diagnosis} was greater than or equal to the sum of all other auxiliary output losses.  Because cross-entropy loss performance deteriorates under scenarios of large class imbalance, this was mitigated by utilizing a focal loss function that is characterized by weighing well-classified examples less \cite{focalloss-2017}.  For a binary classification problem this is formally described as follows: 

\begin{equation} \label{eq:focal_loss}
{FL}(p_t) = -\alpha_t(1-p_t)^\gamma log(p_t)
\end{equation}
where $\gamma \geq 0$ is a focus tuning parameter, $\alpha_t$ is the inverse class frequency tuning parameter, and $p_t$ is defined as follows: 

\begin{equation} \label{eq:pt}
p_t = \begin{cases}
p &\text{if $y=1$}
\\
1-p &\text{otherwise.}
\end{cases}
\end{equation}
In this experiment a focal loss was used for all categorical output targets with parameters $\alpha$ and $\gamma$ initialized to 2. 
Focal loss was compared to cross-entropy loss as a sanity check and performance improvements were observed when using focal loss in regards to both model training time and predictive accuracy.  For  \textit{age} and \textit{density} regression targets, mean squared error (MSE) loss was used.  

Because MAMMO CNN was initialized with ImageNet weights, lower-level CNN features were preserved by using an iterative and stratified training regime motivated by the work of \cite{shen-2017}.  First, the fully-connected layer was trained for 1 epoch with the Adam optimizer and a learning rate of $10^{-3}$.  
Then we cycled between training the top-most dense layers and the convolutional layers using a learning rate of $10^{-4}$ for 5 epochs each, followed by $10^{-5}$ for 10 epochs each.
A batch size of 16 was used to train MAMMO CNN and was the largest that fit within GPU memory constraints.  To bias MAMMO CNN to have the best diagnostic performance, at the end of each training epoch the model with the best AUROC for \textit{diagnosis} was monitored  and saved.

Due to the similarity in network architecture of \textit{Classifier} and \textit{Triage} they were both trained identically. 
Due to the increase in network size and complexity, a batch size of 4 was required to fit within the limits of GPU memory. This required manually balancing the training classes, such that an equal number of positive and negative samples were seen during each batch.  Consider Fig.~\ref{fig:mammocnn}, only the dense layers after concatenation were trained and all other layers were preserved and not updated during back-propagation.  Again, the \textit{Adam} optimizer was used with learning rate initialized to $1e^{-4}$ for 5 epochs then $1e^{-5}$ for 15 epochs.  During training all the previously mentioned augmentations were randomly applied to each input mammogram uniquely to provide the maximum amount of input variation.

All models have been generated, trained, validated and tested using Python, \textit{Keras}, and \textit{TensorFlow} on an Ubuntu Linux 16.04 OS and accelerated using two Nvidia GTX 1080 Ti GPUs with 11GB of memory each.

\subsection{CNN methods}
Below are the methods used to improve the deep learning architecture.

\paragraph{Image augmentations} Though exhaustive search for optimal augmentation parameters was not conducted, several items should be noted.  Horizontal and vertical flips were very important for improving classification performance and generalizability, leading to performance gains of nearly 2\% AUROC.  Because we are using image rotation of up to 20 degrees, the image flips helps the network learn generalized tumor representations.  Augmentations including shear, horizontal shifts, vertical shifts and  rotation all improved mammogram classification.
\paragraph{Dropout} Exhaustive search for an optimal dropout rate was not conducted. Rather, we used a dropout rate of 0.2 as presented in \cite{krysztof-etal-2017} for comparison purposes. Both increasing and decreasing dropout rates were experimented with, but did not conclusively benefit performance.  
\paragraph{Loss function} Although focal loss \cite{focalloss-2017} did not improve classification performance compared to categorical cross-entropy, it did significantly improve training time.  This is due to the fact that focal loss allows the network to focus on incorrectly classified examples, allowing for quicker convergence.
\paragraph{Test time augmentations} We performed test time augmentation at inference time and took the average score over 100 samples.  This significantly helped classification performance (over 3\% in both AUROC and AUPRC). By providing the network with various ``perspectives'' of the same mammogram, test time augmentation mitigated the likelihood of misinterpreting a solitary mammogram.  For a given patient, we tried varying all the mammograms with identical augmentation (across views), but this did not improve performance as much as providing each mammogram view with it's own random augmentation seed, such that each mammogram view will have a unique sequence of random augmentations relative to each other.  For example, a patient's MLO right mammogram may be flipped vertically and rotated 10 degrees, and the patient's corresponding CC left mammogram may be rotated 3 degrees and flipped horizontally.  We also tried to provide all four mammogram views to the \textit{Classifier} without any augmentation, but this did not perform as well as with TTA.
\paragraph{Class balancing} To overcome class imbalance issues, we tried two methods.  The first involved using class weighting.  The second involved training the network with an equal number of positive and negative samples.  The data showed class weighting outperforms manually balancing classes because with larger batch sizes of approximately 16 mammograms there is a high likelihood of a positive sample present in each batch.  For larger networks this was not the case.  When training \textit{Classifier} we were limited to batch sizes of 4, necessitating manual class balancing.
\paragraph{Multi-channel CLAHE augmentation} We used the CLAHE augmentation technique presented in \cite{teare-2017} that encoded various CLAHE clipping and kernel sizes across each RGB color channel.  This worked perfectly into our Inception-Resnet\_V2 architecture, because it was trained using color imagery.  We take their approach a step further and applied augmentation to both the kernel size and clip limit.  This improved network generalizability as well by providing another degree of freedom for image augmentation.  
\paragraph{Gaussian noise} We added Gaussian noise as an augmentation presented in \cite{gaussian-2015} with a $\sigma$ of 0.01.  This improved performance by providing both generalizability and regularization.

\paragraph{Multi-view augmentation} When training \textit{Classifier}, instead of using each mammogram view in it's own respective channel, we tried randomly assigning mammogram views during training and prediction, i.e., each CNN could have as input any of MLO right, MLO left, CC right, or CC left mammogram views.  As far as we tested, this hurt performance, but perhaps could have helped if we had the time to increase our network size or find optimal hyper-parameters.  
\paragraph{Random cropping augmentation} We tried using the random cropping and augmentation scheme used in \cite{krysztof-etal-2017}, but it did not improve performance in comparison to our presented augmentation pool. The level of augmentations they presented did not include aggressive enough rotations and flips.  
\paragraph{External datasets} We translated the CBIS-DDSM annotation to \textit{Tommy} annotations as best as possible and used the optical scale conversions provided by \cite{ddsm-database1, ddsm-database2}.  We added the CBIS-DDSM to our dataset to train the MAMMO CNN, but it did not improve performance significantly, if at all. This is probably due to the fact that CBIS-DDSM is scanned film mammograms and \textit{Tommy} is digital.

\begin{figure}[t]
\centering
  \includegraphics[width=\linewidth]{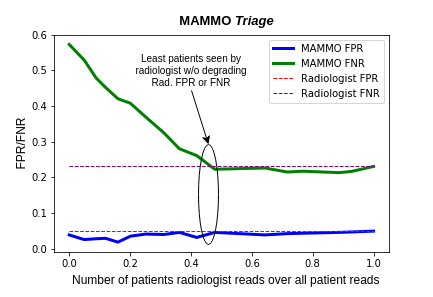}
  \caption{Operating points of MAMMO triage in terms of the false positive rate and false negative rate. }
  \label{fig:MAMMOoperating}
\end{figure}

\section{\label{Appendix:triage} MAMMO Triage}

Fig.~\ref{fig:MAMMOoperating} illustrates MAMMO \textit{Triage} performance on the 1000 patient test set used in Table~\ref{table:final}, and shows the FPR/FNR versus the number of patients the radiologist reads over all patient reads (1000).  This graph can be interpreted as depicting the various MAMMO operating points as patients are partitioned between the \textit{Classifier} and radiologist, where all patients are seen by the \textit{Classifier} at $x = 0.0$ and all patients are read by the radiologist at $x = 1.0$.  The annotated region represents the operating point that satisfied Eq.~\ref{eq:maximize} and presented in Table~\ref{table:final}.

\begin{table}[t!] 
\centering
\caption{ \label{table-tommypatients}\textit{Tommy} patient participant characteristics for age, breast density, and dominant radiological features.  }
\begin{tcolorbox}[tab2,tabularx={l|c|c}]
    {\normalfont \small \bf \textcolor{red!60!black}{Population}} & 
    
    {\normalfont \small \bf \textcolor{red!60!black}{Total (\%) }} &
    {\normalfont \small \bf \textcolor{red!60!black}{Cancer (\%)}} 
    \\ \hline \hline
    {\normalfont \small Age $40-49$ years}   & {\normalfont \small 6} &{\normalfont \small 3} \\ \hline 
    {\normalfont \small  Age $50-59$ years}   &{\normalfont \small 59} &{\normalfont \small 40} \\ \hline
    {\normalfont \small Age $60-69$ years}   & {\normalfont \small 29} &{\normalfont \small 45} \\ \hline 
    {\normalfont \small  Age $\geq70$ years}   &{\normalfont \small 6} &{\normalfont \small 12} \\ \hline \hline
    {\normalfont \small Breast density $0\%-24\%$}   & {\normalfont \small 27} &{\normalfont \small 33} \\ \hline 
    {\normalfont \small   Breast density  $25\%-49\%$}   &{\normalfont \small 43} &{\normalfont \small 38} \\ \hline
    {\normalfont \small  Breast density $50\%-74\%$}   & {\normalfont \small 23} &{\normalfont \small 24} \\ \hline 
    {\normalfont \small   Breast density  $75\%-100\%$}   &{\normalfont \small 7} &{\normalfont \small 5} \\ \hline \hline
    {\normalfont \small Circumscribed mass}   & {\normalfont \small 31} &{\normalfont \small 14} \\ \hline 
    {\normalfont \small  Spiculated mass}   &{\normalfont \small 13} &{\normalfont \small 44} \\ \hline
    {\normalfont \small Micro-calcification}   & {\normalfont \small 17} &{\normalfont \small 24} \\ \hline 
    {\normalfont \small  Distortion}   &{\normalfont \small 8} &{\normalfont \small 9} \\ \hline 
    {\normalfont \small  Asymmetrical distortion}   &{\normalfont \small 31} &{\normalfont \small 9} \\ \hline
\end{tcolorbox}
\end{table}

\section{\label{Appendix:Tommy} Tommy patient distribution}

Table~\ref{table-tommypatients} shows the patient distributions for the \textit{Tommy} dataset across age, breast density, and dominant radiological features.

\section*{Acknowledgment}
The authors would like to thank Auke van der Schaar for providing invaluable recommendations for improving the MAMMO CNN and \textit{Classifier}, as well as our colleagues Kartik Ahuja, Ahmed Alaa, James Jordon, Jinsung Yoon, and the entire UCLA ML-AIM research group for providing insightful comments for improving this work.   


\ifCLASSOPTIONcaptionsoff
  \newpage
\fi


%

\bibliographystyle{IEEEtran.bst}
\bibliography{kyono-mammo}

%








\end{document}